%% file: main.tex
\RequirePackage{filecontents}

\documentclass[9pt,onecolumn,twoside]{srformat_fixed}
\usepackage{stfloats}
\usepackage{float}

\usepackage{epstopdf}
\usepackage{dsfont}
\usepackage{bbm}
\usepackage{filecontents}
\usepackage[style=nature, maxnames=99, backend=biber, bibencoding=utf8]{biblatex}
\let\cite\supercite
\usepackage[normalem]{ulem}
\usepackage{pifont}
\UseRawInputEncoding
\usepackage{makecell}
\usepackage{amssymb}
\usepackage{amsthm}
\usepackage{multirow}
\usepackage{array}
\usepackage{adjustbox} 
\usepackage{wrapfig}
\usepackage{subcaption} 

\usepackage{framed}
\usepackage{bm}
\usepackage{nicefrac}
\usepackage{diagbox}
\usepackage{listings}
\usepackage{multicol}
\usepackage{marvosym}
\usepackage{pdfpages}
\usepackage{appendix}
\usepackage{balance}
\usepackage{longtable}
\usepackage{xurl}
\setlist{nolistsep}

\hypersetup{colorlinks=true, allcolors=blue, pdfborder={0 0 0}}

\input{macros}

\runningtitle{ } 
\runningauthor{Jiang \textit{et al.}}

\title{\modelName: A Hierarchical Reasoning Foundation Model for Scalable and Interpretable Surgical Decision Support with Multi-Center Clinical Validation}

\author[1,$\dagger$]{\textbf{Jian Jiang}}
\author[1,$\dagger$]{\textbf{Chenxi Lin}}
\author[1]{\textbf{Yiming Gu}}
\author[5]{\textbf{Zengyi Qin}}
\author[3]{\textbf{Zhitao Zeng}}
\author[9]{\textbf{Kun Yuan}}
\author[4]{\textbf{Yonghao Long}}
\author[6]{\textbf{Xiang Xia}}
\author[1]{\textbf{Cheng Yuan}}
\author[1]{\textbf{Yuqi Wang}}
\author[8]{\textbf{Zijie Yue}}
\author[1]{\textbf{Kunyi Yang}}

\author[1]{\textbf{Yuting Zhang}}
\author[3]{\textbf{Zhu Zhuo}}
\author[14]{\textbf{Dian Qin}}
\author[7]{\textbf{Xin Wang}}
\author[13]{\textbf{NG Chi Fai}}
\author[5]{\textbf{Brian Anthony}}
\author[12]{\textbf{Daguang Xu}}
\author[10,11]{\textbf{Guy Rosman}}
\author[10]{\textbf{Ozanan Meireles}}
\author[6]{\textbf{Zizhen Zhang}}
\author[9]{\textbf{Nicolas Padoy}}
\author[2]{\textbf{Hesheng Wang}}
\author[4]{\textbf{Qi Dou}}
\author[3]{\textbf{Yueming Jin}}
\author[1,$\ast$]{\textbf{Yutong Ban}}

\affil[1]{Global College, Shanghai Jiao Tong University, Shanghai, China}
\affil[2]{School of Automation and Intelligent Sensing, Shanghai Jiao Tong University}
\affil[3]{Department of Biomedical Engineering and Department of Electrical and Computer Engineering, National University of Singapore, Singapore}
\affil[4]{Department of Computer Science and Engineering, The Chinese University of Hong Kong, Hong Kong SAR, China}
\affil[5]{Institute for Medical Engineering \& Science, Massachusetts Institute of Technology, Cambridge, MA, USA}
\affil[6]{Department of Gastrointestinal Surgery, Renji Hospital, Shanghai Jiao Tong University School of Medicine, Shanghai, China}
\affil[7]{Division of Pancreatic Surgery, Department of General Surgery, West China Hospital of Sichuan University, Chengdu, China}
\affil[8]{College of Electronic and Information Engineering, Tongji University, Shanghai, China}
\affil[9]{ICube, University of Strasbourg, CNRS, IHU Strasbourg, France}
\affil[10]{Massachusetts General Hospital, Massachusetts, US}
\affil[11]{Department of Surgery, Duke University, Durham, NC 27707, USA}
\affil[12]{Nvidia, US}
\affil[13]{Department of Surgery, The Chinese University of Hong Kong, Hong Kong SAR, China}
\affil[14]{Chengdu Withai Innovations Technology Company, Chengdu, China}

\affil[$\dagger$]{These authors contributed equally to this work.}

\correspondingauthoraffiliation[$\ast$]{Corresponding author: Yutong Ban (yban@sjtu.edu.cn)}

\begin{abstract}
Surgical scene understanding demands not only accurate predictions but also interpretable reasoning that surgeons can verify against clinical expertise. However, existing surgical vision-language models generate predictions without reasoning chains, and general-purpose reasoning models fail on compositional surgical tasks without domain-specific knowledge. We present \modelName{}, a surgical Vision-Language Model that addresses this gap through hierarchical reasoning trained via a four-stage pipeline. Our approach introduces three key contributions: (1) a three-level reasoning hierarchy decomposing surgical interpretation into perceptual grounding, relational understanding, and contextual reasoning; (2) the largest surgical chain-of-thought dataset with 320,000 reasoning pairs; and (3) a four-stage training pipeline progressing from supervised fine-tuning to group relative policy optimization and iterative self-improvement. Evaluation on \benchmarkName{}, comprising four public benchmarks and six multi-center external validation datasets from five institutions, demonstrates that \modelName{} achieves the highest Arena Score (57.7\%) on public benchmarks versus Gemini 3.0 Pro (29.8\%) and GPT-5.1 (28.5\%), outperforming both proprietary reasoning models and specialized surgical VLMs on the majority of tasks spanning triplet recognition, phase recognition, action recognition, and critical view of safety assessment, with a 15.2 percentage point improvement over the strongest surgical baseline on external validation.
\end{abstract}

\keywords{Surgical Scene Understanding, Vision-Language Model, Chain-of-Thought Reasoning, Hierarchical Reasoning}

\dates{\rec{xx xx, xxxx} \acc{xx xx, xxxx}}

\begin{document}

\maketitle
\blfootnote{Project page: \url{https://jianjiangkcl.github.io/Surg-R1/}}
\vspace{-13pt}

\section{INTRODUCTION}
\input{1_introduction}

\section{RESULTS}
\input{2_results}

\section{DISCUSSION}
\input{8_discussion}

\phantomsection
\section{METHODS}
\label{sec:methods}
\input{5_method}

\clearpage
\section{SUPPLEMENTARY MATERIALS}
\input{9_supp}

\clearpage
\section{Acknowledgments} 
We thank the clinical teams at West China Hospital, Nanfang Hospital, Renji Hospital, University Hospital of Strasbourg, and the Chinese University of Hong Kong for providing the external validation datasets. We also thank the creators of the public surgical benchmarks used in this study.

\section{Declarations}

\textbf{Funding}
This work has been supported by the program of National Natural Science Foundation of China (No. 62503322) and by Shanghai Magnolia Funding Pujiang Program (No. 23PJ1404400 and No. 24PJD047) 

\textbf{Conflict of interest}
The authors declare no competing interests.

\textbf{Ethics approval and consent to participate}
This study used publicly available surgical datasets that were collected under ethics approvals granted by their respective institutions. External validation data from West China Hospital, Nanfang Hospital, Renji Hospital, University Hospital of Strasbourg, and the Chinese University of Hong Kong were collected under protocols approved by the corresponding institutional review boards. All patient data were fully de-identified prior to use in this study.

\textbf{Consent for publication}
All authors have reviewed the final manuscript and consent to its publication.

\textbf{Data availability}
The training data, evaluation benchmarks, and associated materials will be made publicly available upon publication of this work at the project page: \url{https://jianjiangkcl.github.io/Surg-R1/}.

\textbf{Materials availability}
All the materials that support the findings of this work are presented in the paper.

\textbf{Code availability} 
The source code and model weights will be released upon publication at the project page: \url{https://jianjiangkcl.github.io/Surg-R1/}.

\textbf{Author contribution}
\textbf{Conceptualization:} Jiang, Lin, Ban, and Gu; \textbf{Methodology:} Jiang and Lin; \textbf{Software:} Jiang and Lin; \textbf{Validation:} Jiang, Lin, Gu, and Zeng; \textbf{Formal Analysis:} Jiang, Lin, and Gu; \textbf{Investigation:} Jiang, Lin, Gu, and Zeng; \textbf{Data Curation:} Long, K. Yang, and Xia; \textbf{Writing -- Original Draft:} Jiang, Lin, Gu, C. Yuan, Wang, and Ban; \textbf{Writing -- Review and Editing:} All authors; \textbf{Supervision:} Ban, Jin, and Dou.

\newpage
\printbibliography[title={REFERENCES}]


\end{document}

%% file: macros.tex
\newcommand{\modelName}{Surg-R1}                     
\newcommand{\datasetName}{Surg-R1-Data}              
\newcommand{\benchmarkName}{SurgBench}               

\newcommand{\modelFinal}{\textit{\modelName}}        
\newcommand{\datasetCoT}{\textit{\datasetName}}      
\newcommand{\benchmark}{\textit{\benchmarkName}}     
\newcommand{\datasetBase}{\textit{SurgDB-3M}}        

\newcommand{\ModelLLaVAcite}{LLaVA-1.5-7B\cite{liu2024improved}}

\newcommand{\ModelPhicite}{Phi-4-Multimodal\cite{abouelenin2025phi}}

\newcommand{\ModelMistralFullCite}{Mistral-Small-3.1-24B-Instruct-2503\cite{mistral2025small}}

\newcommand{\ModelInternVLSmallcite}{InternVL3-8B\cite{zhu2025internvl3}}

\newcommand{\ModelInternVLLargeCite}{InternVL3-78B\cite{zhu2025internvl3}}

\newcommand{\ModelMiniCPMVcite}{MiniCPM-V-2\_6\cite{openbmb2024minicpm}}

\newcommand{\ModelMiniCPMOcite}{MiniCPM-O-2\_6\cite{openbmb2024minicpm}}

\newcommand{\ModelGemmacite}{Gemma3-27B-it\cite{team2025gemma}}

\newcommand{\ModelSkyworkCite}{Skywork-R1V-38B\cite{peng2025skywork}}

\newcommand{\ModelLlamaCite}{Llama-4-Scout-17B\cite{meta2025llama4}}
\newcommand{\ModelLlamaFullCite}{Llama-4-Scout-17B-16E-Instruct\cite{meta2025llama4}}

\newcommand{\ModelQwenSmallcite}{Qwen2.5-VL-7B-Instruct\cite{bai2025qwen25vl}}

\newcommand{\ModelQwenMediumcite}{Qwen2.5-VL-32B-Instruct\cite{bai2025qwen25vl}}

\newcommand{\ModelQwenLargecite}{Qwen2.5-VL-72B-Instruct\cite{bai2025qwen25vl}}

\newcommand{\ModelQwenMaxcite}{Qwen2.5-Max\cite{bai2025qwen25vl}}

\newcommand{\ModelGPTFourOFullOCite}{GPT-4o-2024-0806\cite{hurst2024gpt}}

\newcommand{\ModelGPTFiveCite}{GPT-5.1\cite{singh2025openai}}
\newcommand{\ModelGPTFiveThinkCite}{GPT-5.1-Thinking\cite{singh2025openai}}

\newcommand{\ModelGeminiFlashCite}{Gemini-2.0-Flash\cite{team2023gemini}}

\newcommand{\ModelGeminiProCite}{Gemini-3.0-Pro\cite{google2025gemini3}}
\newcommand{\ModelGeminiProThinkCite}{Gemini-3.0-Pro-Thinking\cite{google2025gemini3}}

\newcommand{\ModelQwenSurgCite}{Qwen2.5-VL-7B-Surg-CholecT50\cite{nvidia2025surg}}

\newcommand{\SegColCite}{SegCol\cite{ju2024segcol}}
\newcommand{\CholecEightyCite}{Cholec80\cite{twinanda2016endonet}}
\newcommand{\CholecEightyCVSCite}{Cholec80-CVS\cite{rios2023cholec80}}
\newcommand{\CholecFiftyCite}{CholecT50\cite{nwoye2022rendezvous}}
\newcommand{\AutoLaparoCite}{AutoLaparo\cite{wang2022autolaparo}}
\newcommand{\CholecEightyVQACite}{Cholec80-VQA\cite{seenivasan2023surgicalgpt}}
\newcommand{\PSIAVAVQACite}{PSI-AVA-VQA\cite{seenivasan2023surgicalgpt}}
\newcommand{\EndoVisEighteenVQACite}{EndoVis18-VQA\cite{seenivasan2023surgicalgpt}}
\newcommand{\CoPESDCite}{CoPESD\cite{wang2025copesd}}
\newcommand{\DSADCite}{DSAD\cite{carstens2023dresden}}
\newcommand{\EndoscapesCite}{Endoscape\cite{murali2023endoscapes}}
\newcommand{\GraSPCite}{GraSP\cite{ayobi2025pixel}}
\newcommand{\MultiBypassCite}{MultiBypass140\cite{lavanchy2024challenges}}
\newcommand{\PitVQACite}{PitVQA\cite{he2024pitvqa}}
\newcommand{\CholecInstanceSegCite}{CholecInstanceSeg\cite{alabi2025cholecinstanceseg}}
\newcommand{\CholecTrackCite}{CholecTrack20\cite{nwoye2023cholectrack20}}
\newcommand{\SARRARPCite}{SAR-RARP50\cite{psychogyios2023sar}}
\newcommand{\EndoVisSeventeenCite}{EndoVis17\cite{allan20192017}}



%% file: 1_introduction.tex
Foundation models have transformed biomedical artificial intelligence in radiology\cite{liu2023radiologyllama2bestinclasslargelanguage, wu2025towards, paschali2025foundation,shui2025largescale}, pathology\cite{xu2025versatile, chen2024towards, wang2024pathology, lu2024visual}, and molecular biology\cite{xu2024smiles, shoshan2024mammal, cheng2023accurate}. However, surgery remains comparatively underexplored\cite{maierhein2022surgical} despite its clinical importance. Intraoperative scenes present distinct challenges: anatomy shifts dynamically throughout procedures, safety-critical decisions carry irreversible consequences, and visual conditions are degraded by poor tissue contrast, variable lighting, and frequent obstructions from blood, smoke, or instruments. These factors constrain direct transfer of approaches developed for static imaging modalities.

Early surgical scene understanding relied on convolutional networks or compact transformer architectures trained for individual tasks\cite{seenivasanSurgicalVQAVisualQuestion2022, hePitVQAImageGroundedText2024}, achieving reasonable accuracy on specific benchmarks but with limited cross-task generalization. Recent surgical Vision-Language Models (VLMs)\cite{zengSurgVLMLargeVisionLanguage2025, jin2024surgicalllavasurgicalscenariounderstanding, wang2025surgicallvlmlearningadaptlarge, shui2025largescale} and Video-Language Models (VideoLMs)\cite{WanZip_CholecMamba_MICCAI2025, wang2026surgvidlmmultigrainedsurgicalvideo,Yua_HecVL_MICCAI2024} fine-tune multimodal foundation models to handle tool recognition, phase identification, and visual question answering within a unified framework, yet generate predictions without explicit reasoning chains. Surgeons need interpretable rationales they can verify against their expertise and incorporate into clinical decision-making.

Several works attempt to incorporate structured reasoning into surgical VLMs. Multi-agent frameworks and instruction tuning with curated explanations can reduce hallucinations, but each has practical limitations. API-dependent agent frameworks\cite{lowSurgRAWMultiAgentWorkflow2025,ZhaJie_CSAPAssist_MICCAI2025} cannot be trained end-to-end and incur high inference costs; locally deployed multi-agent collaboration\cite{low2025carescollaborativeagenticreasoning} provides insufficient coordination between reasoning, perception, and clinical knowledge; schema-structured annotation for instruction tuning\cite{li2024llavasurgmultimodalsurgicalassistant} does not scale. Recent datasets, including SurgVLM-DB\cite{zengSurgVLMLargeVisionLanguage2025}, Surg-396K\cite{wangEndoChatGroundedMultimodal2025}, and EndoVQA-Instruct\cite{liu2025comprehensive} expand data volume but use VQA pairs or unstructured captions rather than structured reasoning annotations. These limitations highlight the need for scalable approaches that can learn structured reasoning without extensive manual annotation.

Reinforcement learning for language models offers a promising direction. DeepSeek-R1\cite{deepseekai2025deepseekr1incentivizingreasoningcapability} shows that, after chain-of-thought (CoT)\cite{CoT_wei2022chain} cold-start, Group Relative Policy Optimization (GRPO)\cite{shao2024deepseekmath} can induce latent reasoning using outcome-based rewards, significantly reducing the reliance on dense process supervision. Current RL-based reasoning approaches across medical and surgical domains vary in CoT supervision and reasoning structure. Without structured CoT supervision, outcome-driven approaches apply GRPO with format and correctness rewards to elicit rationales\cite{PanJia_MedVLMR1_MICCAI2025,XuHui_MedGroundR1_MICCAI2025,su2025medgrpomultitaskreinforcementlearning}, but the resulting reasoning trajectories are less controllable and harder to evaluate clinically. To impose stronger controllability and reasoning consistency, recent methods incorporate explicit CoT cold-start supervision prior to GRPO. For instance, Surgery-R1\cite{hao2025surgeryr1advancingsurgicalvqlareasoning} and SureonVLM-R1\cite{perez2026sureonbenchmarkvisionlanguagemodelsurgical} integrate step templates into CoT annotations, while MedScope\cite{li2026medscope} leverages CoT-guided tool-calling for coarse-to-fine video temporal grounding. However, MedScope's multi-round environment interactions incur prohibitive inference latency, precluding real-time intraoperative deployment. Furthermore, both paradigms rely entirely on static training curricula, lacking iterative self-evolution mechanisms needed to continuously generalize beyond their offline-generated seed annotations.

\begin{figure*}[!t]
\centering
\includegraphics[width=\textwidth]{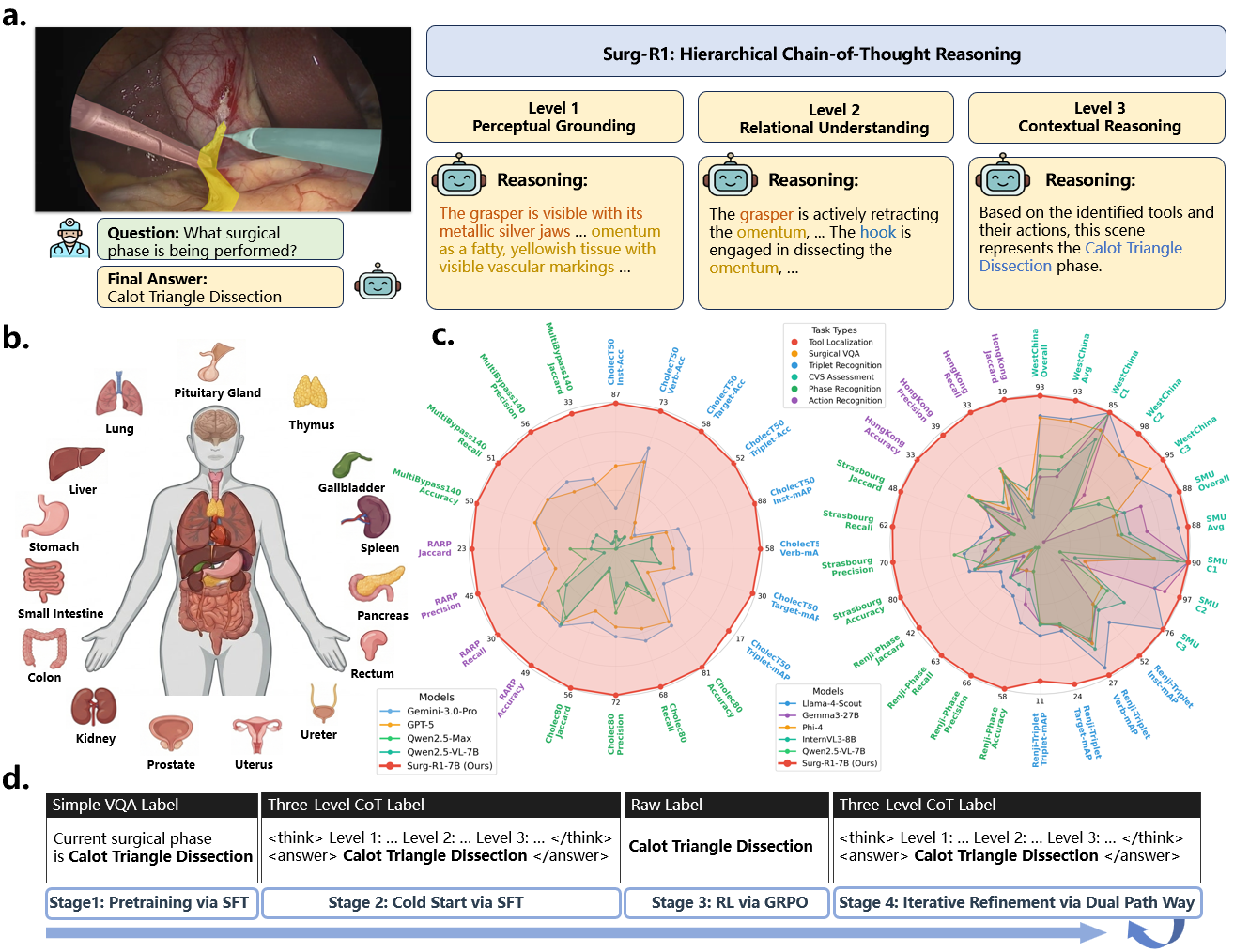}
\caption{{\bf \modelFinal{} framework and performance.} {\bf (a)} Three-level chain-of-thought reasoning. Level 1 identifies instruments and tissues; Level 2 analyzes tool-tissue-action relationships; Level 3 synthesizes contextual reasoning for phase recognition. {\bf (b)} Pre-training dataset (\datasetBase{}) coverage across multiple organ systems and procedure types. {\bf (c)} Performance on public benchmark datasets (left radar) and external validation datasets (right radar) across triplet recognition, phase recognition, action recognition, and critical view of safety (CVS) assessment. {\bf (d)} Four-stage training pipeline progressing from supervised fine-tuning (Stage 1) through chain-of-thought cold-start training (Stage 2) and Group Relative Policy Optimization (Stage 3) to iterative refinement (Stage 4).}
\label{fig:pipeline}
\end{figure*}

We present \modelFinal, a surgical VLM with hierarchical reasoning trained through a four-stage pipeline (Fig.~\ref{fig:pipeline}d). The approach has four components. First, a three-level reasoning hierarchy (perceptual grounding, relational understanding, and contextual reasoning) structures surgical scene interpretation. Second, a surgery-aware CoT synthesis pipeline incorporates domain-specific constraints, including tool and tissue characteristics as well as action vocabularies, and combines ground-truth labels with visual observations to generate forward-reasoning CoT through hierarchical visual grounding rather than reverse-engineering from answers. Third, GRPO with high-entropy token focusing\cite{wang20258020rulehighentropyminority} concentrates learning on decision-critical tokens where the model is most uncertain, enabling scalable reasoning refinement using only final-answer labels as reward signals without requiring CoT annotations. Fourth, an iterative refinement stage expands coverage through rejection sampling and teacher-guided distillation.

Through this pipeline, we construct \datasetCoT{} with approximately 320,000 chain-of-thought pairs and establish \benchmark{}, a comprehensive benchmark spanning ten datasets across four public benchmarks and six multi-center external validation datasets. Evaluation against GPT-5.1, Gemini-3-Pro, and surgical-domain VLMs demonstrates that \modelFinal{} attains the highest Arena Score on both public benchmarks and multi-center external validation across four core surgical tasks, while providing interpretable reasoning outputs aligned with clinical decision-making.

%% file: 2_results.tex
\subsection{Dataset and Experimental Design}

To construct and comprehensively evaluate \modelFinal{}, we curated a large-scale and highly diverse multimodal surgical dataset. To prevent data leakage and ensure rigorous validation, strict isolation was maintained between the training data and the evaluation benchmarks.

Model development was partitioned into foundational pre-training and reasoning-based post-training. For pre-training, we constructed \datasetBase{}, aggregating 2,755,141 image frames across 23 public datasets to establish the robust visual-semantic alignment required for diverse surgical scenes. For post-training, we constructed \datasetCoT{}, a hierarchical chain-of-thought dataset of 320,477 structured reasoning conversations sourced from a representative subset (\CholecEightyCite{}, \CholecFiftyCite{}, \MultiBypassCite{}, \SARRARPCite{}, \EndoscapesCite{}, \EndoVisEighteenVQACite{}, and \EndoVisSeventeenCite{}). Comprising 40,875 GPT-generated and 279,602 self-refined conversations, \datasetCoT{} instills the hierarchical cognitive capabilities necessary for complex clinical inference. Detailed data synthesis protocols are provided in the \hyperref[sec:methods]{Methods}.

\begin{figure*}[htbp]
\centering
\includegraphics[width=\textwidth]{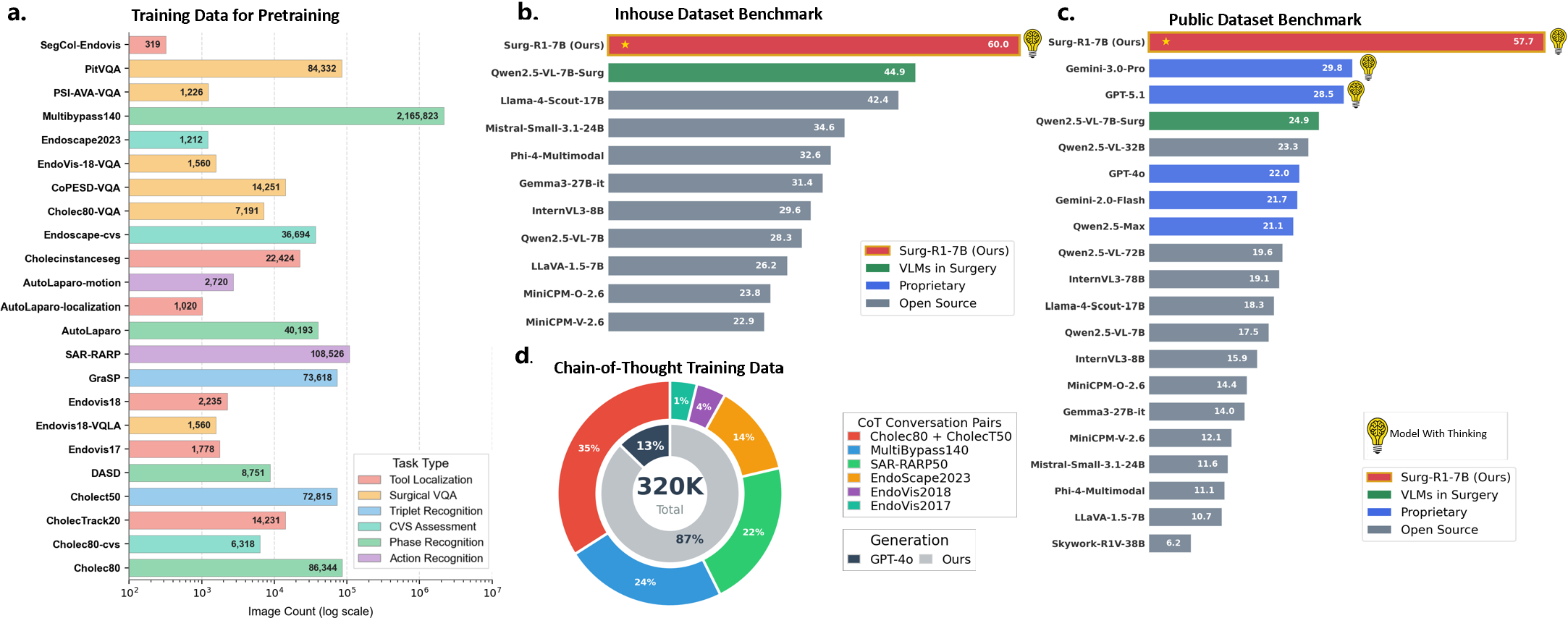}
\caption{{\bf Pre-training data, CoT data composition, and benchmark performance.} {\bf (a)} Pre-training dataset coverage. \datasetBase{} spans 23 surgical datasets covering six task types including tool localization, surgical VQA, triplet recognition, CVS assessment, phase recognition, and action recognition. {\bf (b)} Multi-center external validation datasets. \modelFinal{} achieves the highest arena score, outperforming all baseline models including surgical-domain and proprietary VLMs. {\bf (c)} Public dataset benchmark. \modelFinal{} achieves the highest arena score among all tested models, including reasoning-enhanced models GPT-5.1 and Gemini-3.0-Pro. Models with chain-of-thought capability are indicated by the lightbulb icon. {\bf (d)} Chain-of-thought data composition. \datasetCoT{} comprises CoT conversation pairs across seven datasets, generated through surgery-aware CoT synthesis and model self-improvement.}
\label{fig:arena_scores}
\end{figure*}

We evaluate \modelFinal{} on \benchmark{}, comprising four public benchmarks and six external validation datasets (6,145 unique frames) from five institutions (West China Hospital, Nanfang Hospital, Renji Hospital, University Hospital of Strasbourg, and the Chinese University of Hong Kong). The evaluation spans four core surgical AI tasks: triplet recognition, phase recognition, action recognition, and critical view of safety (CVS) assessment. Baselines include proprietary reasoning-enhanced VLMs (\ModelGPTFiveCite{}, \ModelGeminiProCite{}), general-purpose open-source VLMs (Qwen2.5-VL series\cite{bai2025qwen25vl}, \ModelLLaVAcite{}, \ModelInternVLSmallcite{}, \ModelMiniCPMVcite{}, \ModelPhicite{}, \ModelLlamaCite{}), and the surgical-domain \ModelQwenSurgCite{}. Our three-level reasoning hierarchy (Supplementary Figure~\ref{fig:cot_hierarchy}) structures evaluation from perceptual grounding (Level~1) through relational understanding (Level~2) to contextual reasoning (Level~3), mirroring clinical cognitive processes. Figure~\ref{fig:arena_scores}b,c summarizes the Arena Score across external validation and public benchmark datasets, respectively.

\subsection{Outperforming General-Purpose and Specialized Models}

On public benchmarks, \modelFinal{} achieves an Arena Score of 57.7\%, compared with 29.8\% for Gemini 3.0 Pro and 28.5\% for GPT-5.1 (Figure~\ref{fig:arena_scores}c). Performance advantages are consistent across all task categories, with the largest gains on tasks requiring compositional reasoning.

\subsubsection{Perceptual Grounding Tasks.}
Accurate scene understanding begins with robust perceptual grounding, identifying and localizing discrete anatomical structures and instruments (Level 1 reasoning) despite severe visual degradation such as specular highlights, partial occlusions, and blood obscuration.

\begin{figure}[!htb]
\centering
\includegraphics[width=0.78\linewidth]{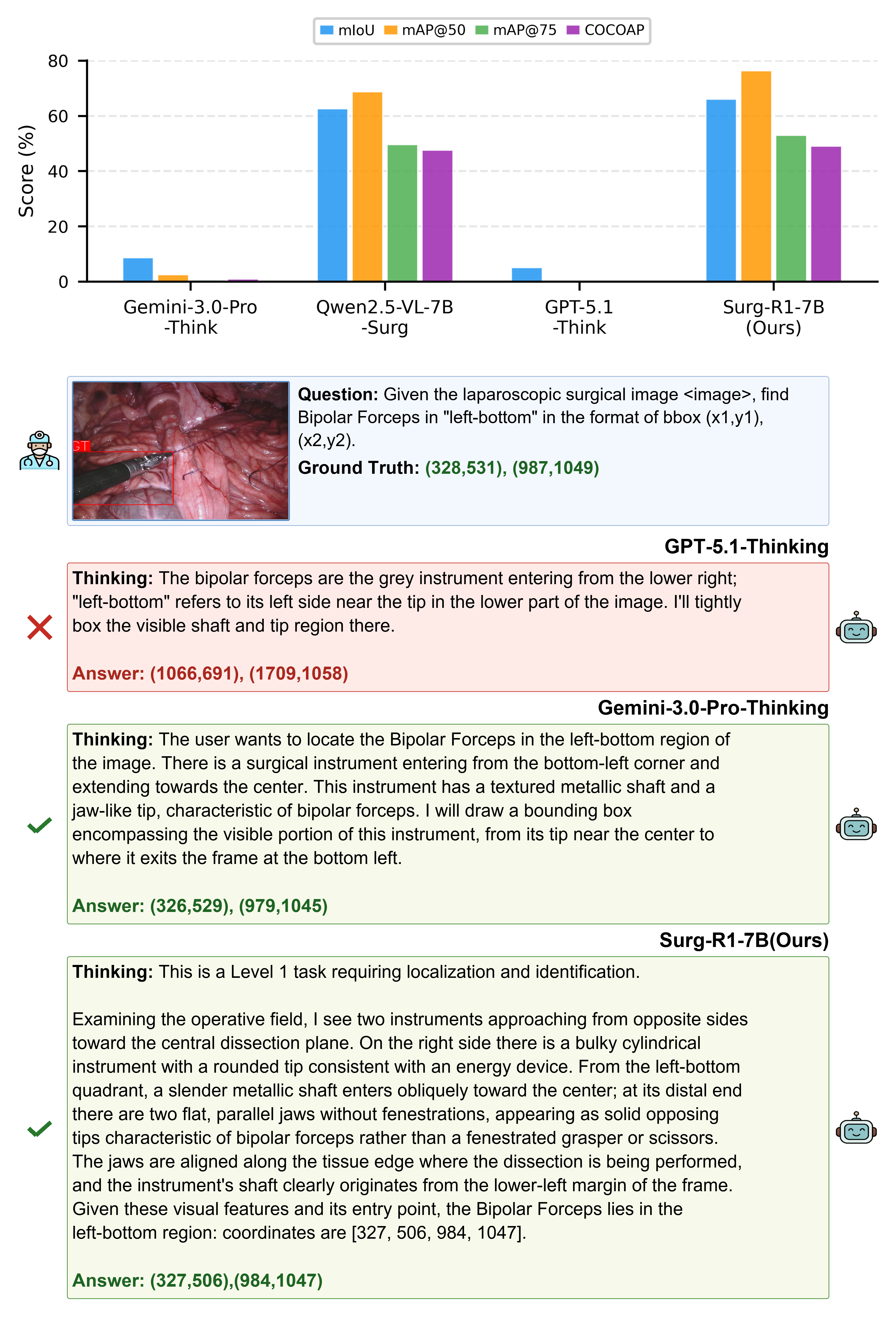}
\caption{{\bf Instrument localization in robotic surgery.} {\bf a.} Performance comparison across models using mIoU, mAP@50, mAP@75, and COCOAP metrics. {\bf b.} Example of Bipolar Forceps localization. GPT-5.1 misidentifies the target instrument, predicting a bounding box in the wrong region (1066,691)-(1709,1058). Gemini 3 correctly localizes the instrument to (326,529)-(979,1045). \modelFinal{} correctly identifies the bipolar forceps through Level 1 reasoning, recognizing ``two flat, parallel jaws without fenestrations'' characteristic of bipolar forceps, producing bounding box (327,506)-(984,1047) that closely matches ground truth (328,531)-(987,1049).}
\label{fig:endovis2017_performance}
\end{figure}

As demonstrated in Figure~\ref{fig:endovis2017_performance}, instrument localization in robotic surgery presents a particularly challenging perceptual grounding task. When asked to localize a bipolar forceps in a laparoscopic scene, GPT-5.1 misidentifies the target instrument entirely, predicting a bounding box in the right side of the frame (1066,691)-(1709,1058) instead of the correct left-bottom region. This failure reflects a fundamental limitation: without domain-specific physical grounding, general VLMs cannot reliably distinguish instruments that share similar metallic appearances. In contrast, \modelFinal{} correctly identifies the bipolar forceps through Level 1 reasoning, explicitly differentiating ``two flat, parallel jaws without fenestrations'' from the energy device on the opposite side, producing bounding box (327,506)-(984,1047) that closely matches the ground truth (328,531)-(987,1049). By reasoning about material properties and morphological features, \modelFinal{} grounds visual features to semantic instrument categories through interpretable hierarchical reasoning, establishing a reliable foundation for downstream compositional tasks.

\subsubsection{Relational Understanding Tasks.}
Building upon basic perception, relational understanding (Level 2 reasoning) analyzes the dynamic physical interactions between identified tools and tissues.

\begin{figure}[!htb]
\centering
\includegraphics[width=0.78\linewidth]{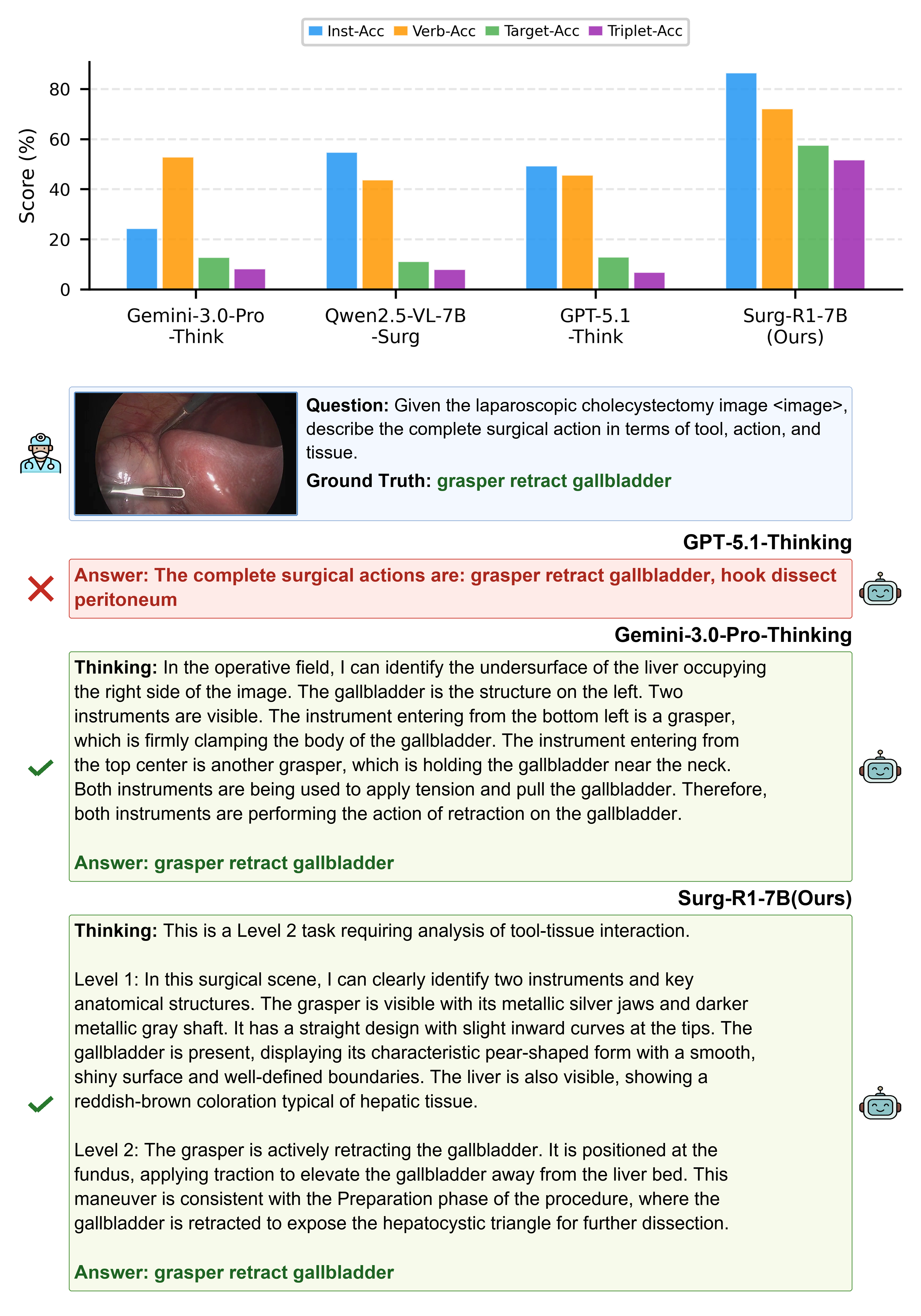}
\caption{{\bf Triplet recognition performance on CholecT50.} {\bf a.} Comparison of instrument, verb, target, and triplet accuracy/mAP across models. {\bf b.} Example of triplet recognition. GPT-5.1 hallucinates an additional action, predicting ``grasper retract gallbladder, hook dissect peritoneum'' when only one triplet is present. Gemini 3 correctly identifies the triplet. \modelFinal{} applies hierarchical reasoning. Level 1 identifies ``metallic silver jaws'' of the grasper and ``pear-shaped form with smooth, shiny surface'' of the gallbladder; Level 2 analyzes the retracting action ``applying traction to elevate the gallbladder away from the liver bed,'' correctly synthesizing (grasper, retract, gallbladder).}
\label{fig:cholect50_performance}
\end{figure}

On the triplet recognition dataset \CholecFiftyCite{} for laparoscopic cholecystectomy, \modelFinal{} achieves 51.69\% triplet accuracy, a 44.9 percentage point improvement over GPT-5.1 (6.77\%; Table~\ref{tab:reasoning_comparison}). This substantial gap exposes the ``binding problem'' inherent in general-purpose chain-of-thought reasoning, where models like GPT-5.1 frequently hallucinate interactions simply because an instrument and tissue are spatially adjacent (Figure~\ref{fig:cholect50_performance}b). \modelFinal{} mitigates this by enforcing physical dependency checks. Before asserting a `retract' action, it explicitly verifies relational cues such as the instrument ``applying traction to elevate the gallbladder,'' ensuring that verbs are physically grounded in observable tool-tissue tension rather than statistical co-occurrence.

\begin{figure}[!htb]
\centering
\includegraphics[width=0.78\linewidth]{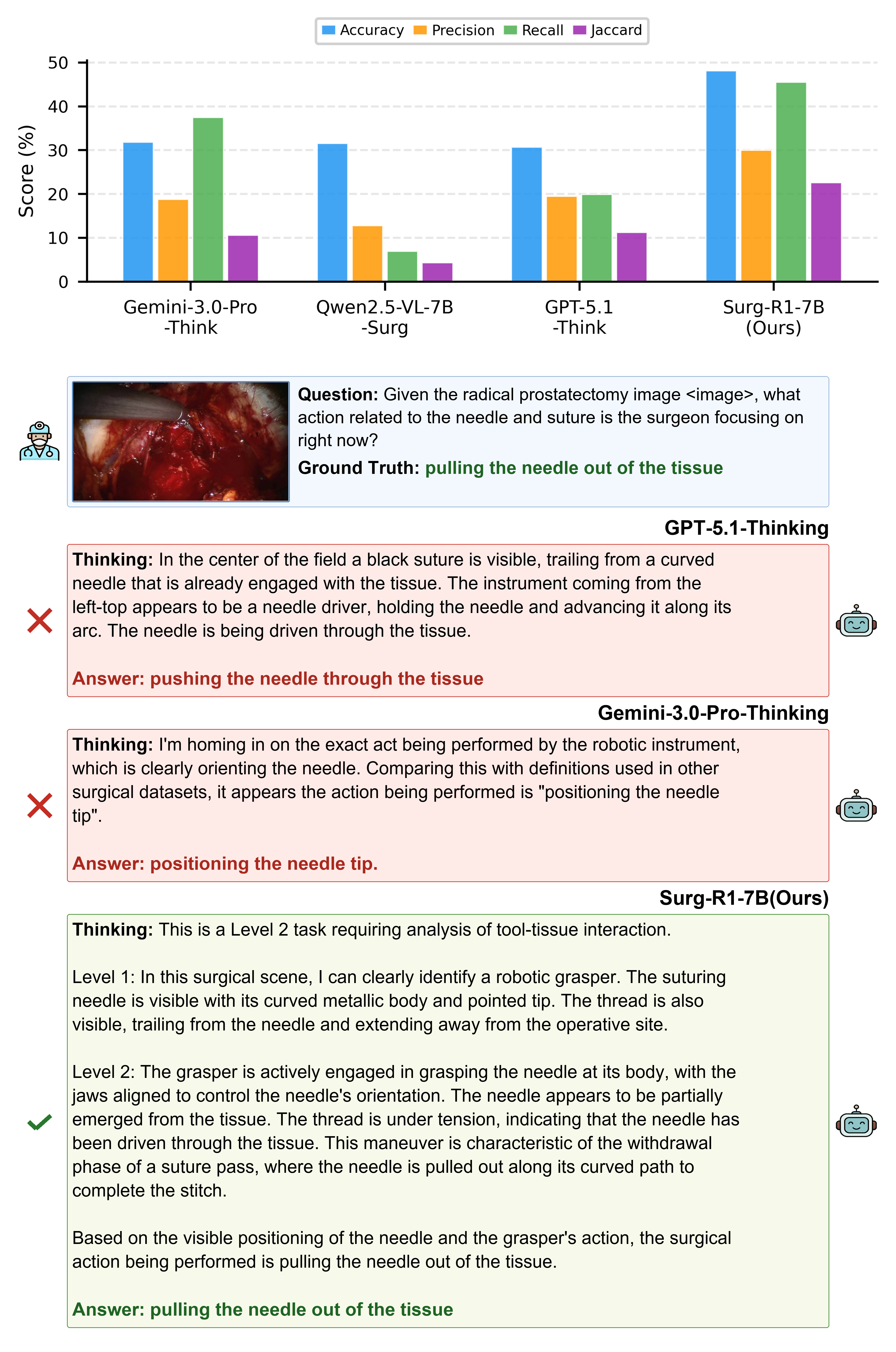}
\caption{{\bf Action recognition on SAR-RARP50 robotic surgery dataset.} {\bf a.} Performance comparison using accuracy, precision, recall, and F1 score. {\bf b.} Example of needle manipulation action recognition. GPT-5.1 incorrectly predicts ``pushing the needle through the tissue,'' misinterpreting the withdrawal phase as insertion. Gemini 3 predicts ``positioning the needle tip,'' failing to recognize the extraction phase. \modelFinal{} correctly identifies ``pulling the needle out of the tissue'' through hierarchical reasoning. Level 1 identifies ``robotic grasper'' and ``suturing needle with curved metallic body''; Level 2 recognizes ``needle appears to be partially emerged from the tissue'' with ``thread under tension,'' characteristic of the ``withdrawal phase of a suture pass.''}
\label{fig:sarrarp_performance}
\end{figure}

This relational capability is further validated in fine-grained action recognition during robotic-assisted radical prostatectomy on the \SARRARPCite{} dataset, where \modelFinal{} achieves 48.10\% accuracy, outperforming Qwen2.5-VL-7B-Surg by 16.6 percentage points. Fine surgical maneuvers are highly susceptible to temporal visual illusions in static frames. For instance, needle insertion and withdrawal share nearly identical spatial configurations (Figure~\ref{fig:sarrarp_performance}b). By analyzing specific relational states (e.g., the suture thread being ``under tension'' and the needle ``partially emerged''), \modelFinal{} effectively disambiguates these kinematic phases without relying on explicit temporal video inputs, outperforming baselines that merely guess based on tool presence.

\subsubsection{Contextual Reasoning Tasks.}
The highest cognitive level involves synthesizing local perceptual and relational data into overarching clinical context, capturing the surgeon's intent (Level 3 reasoning).

\begin{figure}[!htb]
\centering
\includegraphics[width=0.74\linewidth]{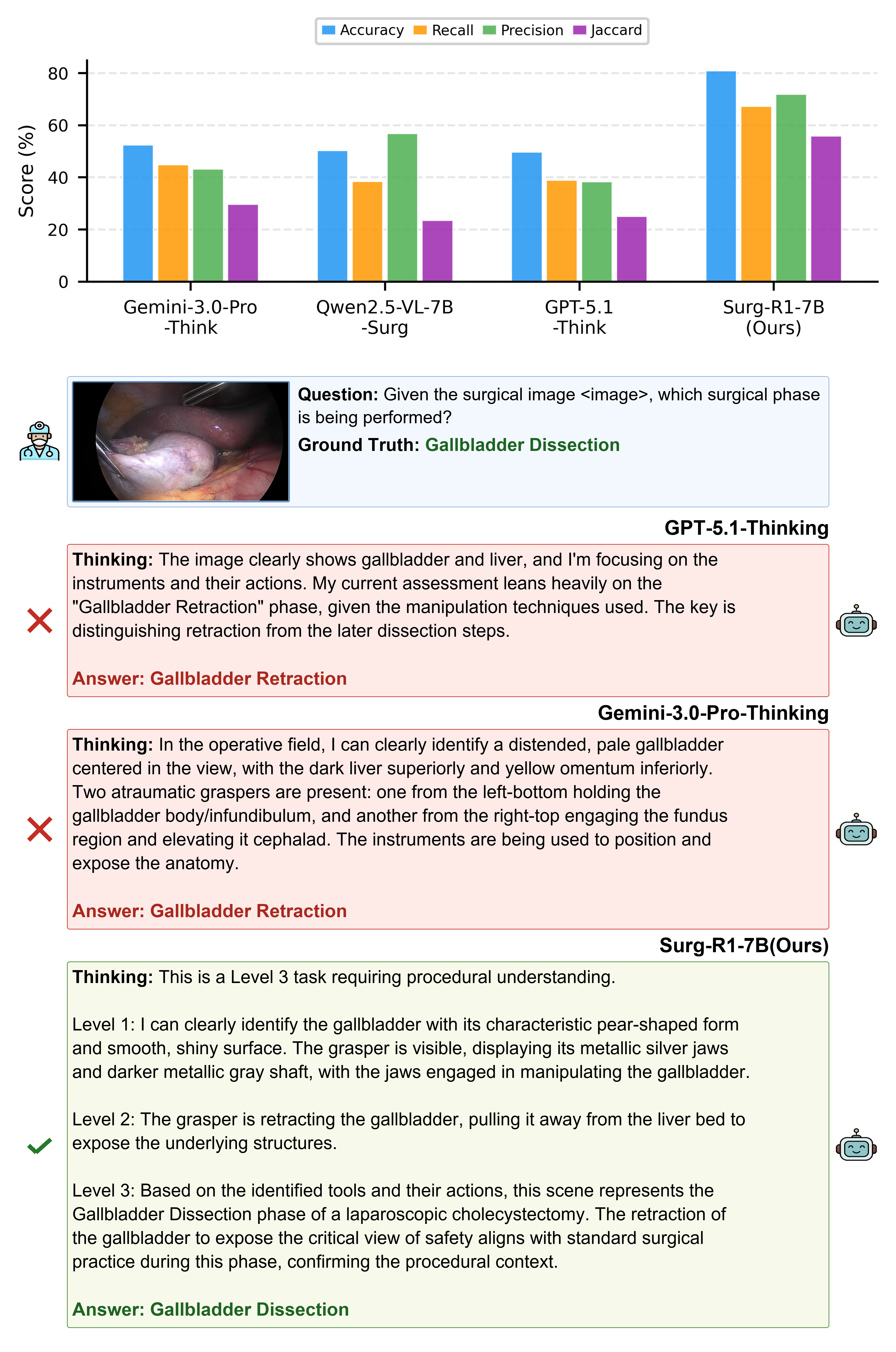}
\caption{{\bf Phase recognition performance on Cholec80.} {\bf a.} Performance comparison using accuracy, recall, precision, and Jaccard index. {\bf b.} Example of phase recognition. Both GPT-5.1 and Gemini 3 incorrectly predict ``Gallbladder Retraction,'' confusing retraction with dissection despite visible tissue manipulation. \modelFinal{} correctly identifies ``Gallbladder Dissection'' through three-level reasoning. Level 1 recognizes the gallbladder and grasper; Level 2 analyzes ``grasper retracting the gallbladder, pulling it away from the liver bed''; Level 3 synthesizes that this action ``exposes the critical view of safety,'' characteristic of the dissection phase rather than simple retraction.}
\label{fig:cholec80_phase}
\end{figure}

On the phase recognition benchmark \CholecEightyCite{}, \modelFinal{} achieves 80.90\% accuracy, outperforming Gemini 3.0 Pro by 28.5 percentage points (Figure~\ref{fig:cholec80_phase}a). Surgical phases are defined by clinical objectives rather than isolated actions. As demonstrated in Figure~\ref{fig:cholec80_phase}b, general models frequently confuse the ``Dissection'' phase with mere ``Retraction'' because they myopically focus on the local action of pulling the tissue. \modelFinal{} avoids this semantic trap by utilizing contextual reasoning, recognizing that the local retraction action is actively ``exposing the critical view of safety,'' thereby synthesizing the true surgical intent (Dissection).

\begin{figure}[!htb]
\centering
\includegraphics[width=0.76\linewidth]{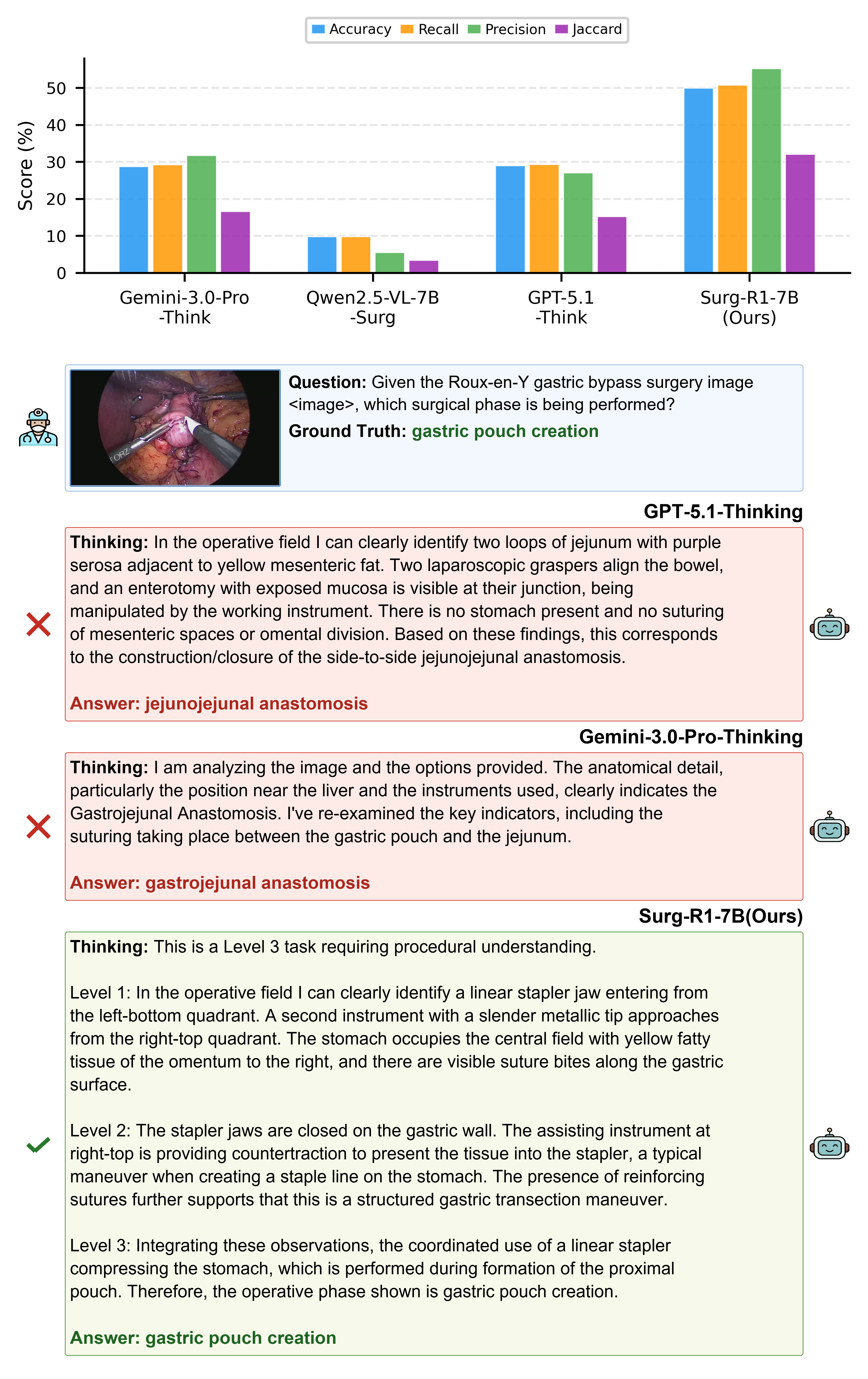}
\caption{{\bf Phase recognition performance on MultiBypass140.} {\bf a.} Cross-procedure generalization evaluation on gastric bypass surgery. {\bf b.} Example of phase recognition in Roux-en-Y gastric bypass. GPT-5.1 incorrectly predicts ``jejunojejunal anastomosis,'' misinterpreting the scene as bowel reconnection. Gemini 3 predicts ``gastrojejunal anastomosis,'' confusing stomach-to-bowel connection with pouch formation. \modelFinal{} correctly identifies ``gastric pouch creation'' through hierarchical reasoning. Level 1 identifies ``linear stapler jaw'' and ``visible suture bites along the gastric surface''; Level 2 recognizes ``stapler jaws closed on gastric wall'' with ``countertraction to present tissue''; Level 3 integrates these as ``formation of the proximal pouch.''}
\label{fig:multibypass_performance}
\end{figure}

This contextual robustness extends across surgical specialties. On \MultiBypassCite{}, \modelFinal{} achieves 49.94\% accuracy versus 29.01\% for GPT-5.1 (Figure~\ref{fig:multibypass_performance}a). In complex procedures like Roux-en-Y gastric bypass, local actions (e.g., a stapler closing) can occur in multiple distinct phases (pouch creation vs. anastomosis). By integrating Level 1 tool identification (stapler, visible sutures) with Level 2 relational interactions (countertraction), the Level 3 reasoning mechanism accurately maps these elements to the global workflow timeline (formation of the proximal pouch; Figure~\ref{fig:multibypass_performance}b), demonstrating that the hierarchical reasoning architecture generalizes beyond single-procedure heuristics.

\subsection{Generalizing Across Clinical Centres}

To comprehensively validate the model's resilience against institutional domain shifts, we evaluated \modelFinal{} on six multi-center external validation datasets from five institutions. As shown in Figure~\ref{fig:arena_scores}b, \modelFinal{} achieves 60.0\% average Arena Score versus 44.9\% for Qwen2.5-VL-7B-Surg, a 15.2 percentage point improvement. This indicates that hierarchical reasoning effectively mitigates domain shifts (such as variations in endoscopic color temperature, camera angles, and institutional surgical habits) that typically degrade end-to-end deep learning models.

\subsubsection{Robust Perceptual Grounding Across Demographics.}
\begin{figure}[!htb]
\centering
\includegraphics[width=0.78\linewidth]{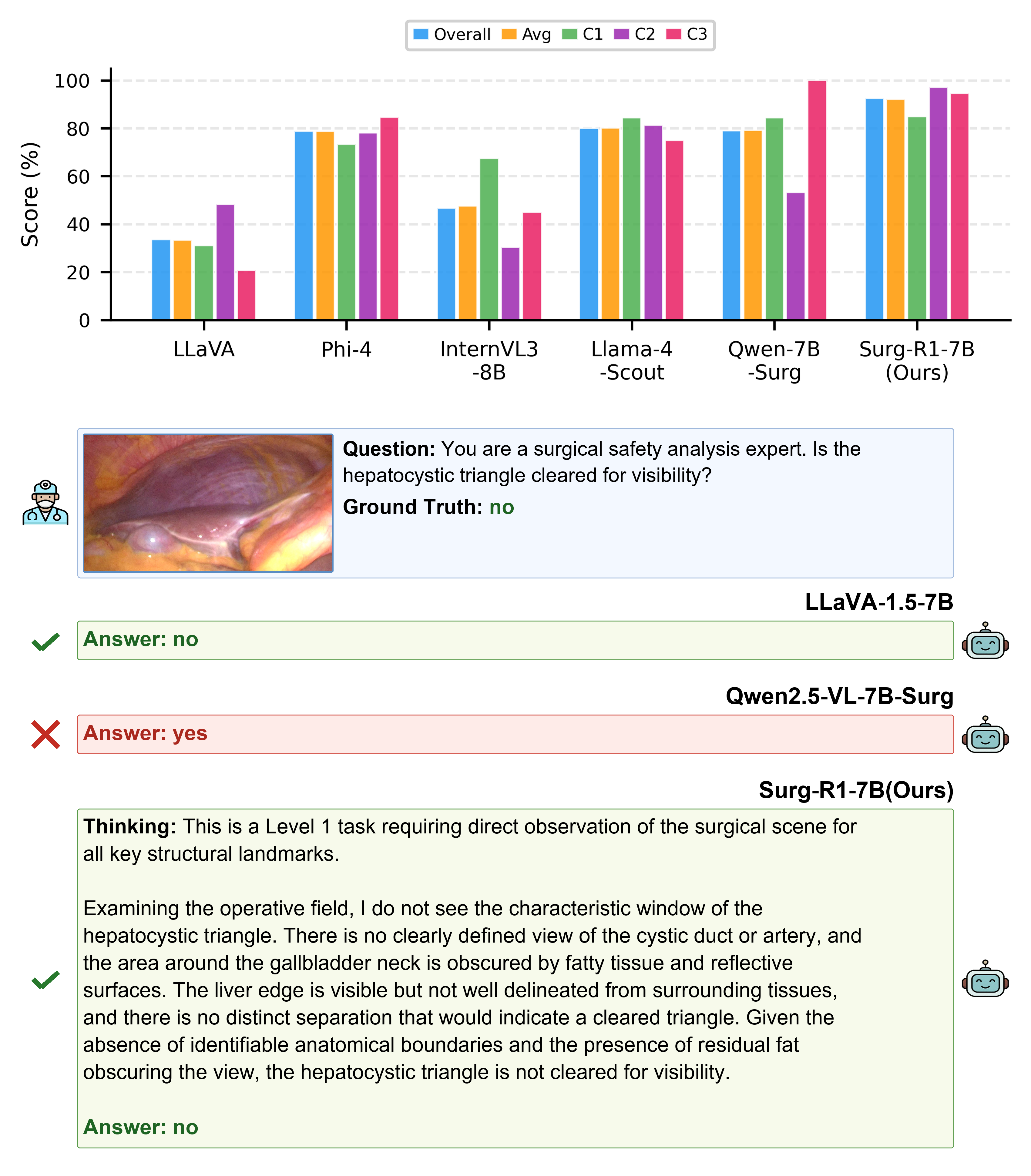}
\caption{{\bf CVS assessment on West China Hospital clinical dataset.} {\bf a.} Performance showing overall accuracy and per-criterion results (C1, C2, C3). {\bf b.} Example assessing hepatocystic triangle clearance. LLaVA-1.5-7B correctly answers ``no.'' Qwen2.5-VL-7B-Surg incorrectly answers ``yes,'' failing to recognize incomplete clearance. \modelFinal{} correctly identifies ``no'' with detailed reasoning, observing ``no clearly defined view of the cystic duct or artery,'' noting ``area around gallbladder neck is obscured by fatty tissue and reflective surfaces,'' and concluding ``no distinct separation that would indicate a cleared triangle.''}
\label{fig:westchina_cvs_performance}
\end{figure}

\begin{figure}[!htb]
\centering
\includegraphics[width=0.78\linewidth]{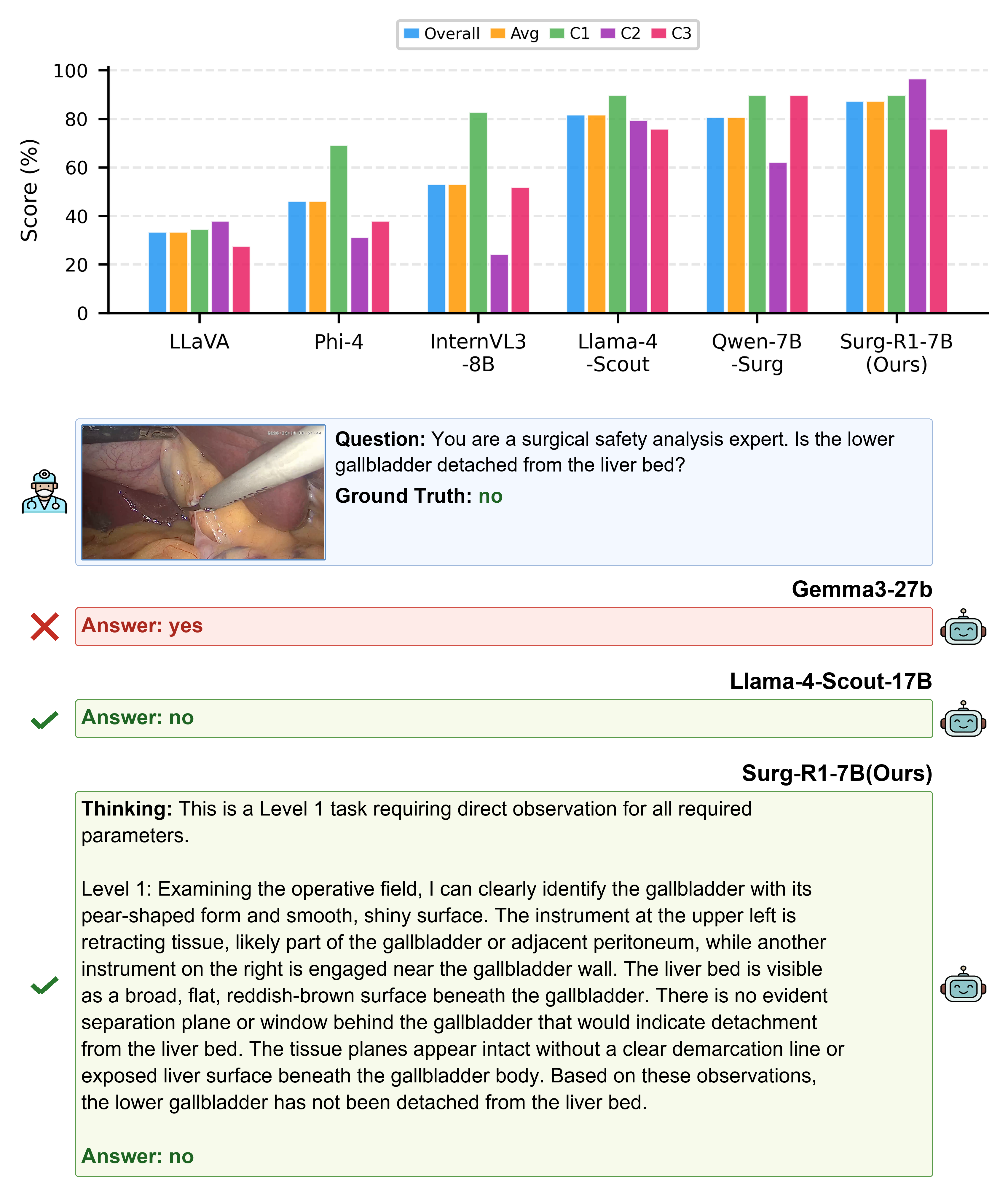}
\caption{{\bf CVS assessment on Nanfang Hospital clinical dataset.} {\bf a.} Performance comparison across models. {\bf b.} Example assessing gallbladder detachment from liver bed. Gemma3-27b incorrectly answers ``yes,'' misinterpreting tissue positioning. Llama-4-Scout-17B correctly answers ``no.'' \modelFinal{} provides accurate assessment with detailed reasoning. Level 1 identifies ``liver bed visible as a broad, flat, reddish-brown surface beneath the gallbladder''; observes ``no evident separation plane or window behind the gallbladder''; concludes ``tissue planes appear intact without a clear demarcation line,'' confirming the gallbladder ``has not been detached from the liver bed.''}
\label{fig:smu_cvs_performance}
\end{figure}

The model's advantage is most pronounced in CVS assessment on external clinical data, where anatomical variations are substantially higher than on public benchmarks. At West China Hospital, \modelFinal{} achieves 92.50\% overall accuracy, and at Nanfang Hospital, it maintains 87.36\% accuracy (Supplementary Tables~\ref{tab:westchina_results},~\ref{tab:smu_results}). When assessing hepatocystic clearance under varying illumination and tissue adiposity (Figures~\ref{fig:westchina_cvs_performance},~\ref{fig:smu_cvs_performance}), baselines often misinterpret shadowing or dense fibrofatty tissue as a cleared window. \modelFinal{} maintains accuracy by anchoring its perceptual grounding to structural invariant features, explicitly noting the lack of a ``distinct separation plane'' or the presence of obscuring ``fatty tissue,'' ensuring safety assessments remain conservative and reliable across diverse patient demographics.

\subsubsection{Invariant Relational Understanding.}
\begin{figure}[!htb]
\centering
\includegraphics[width=0.78\linewidth]{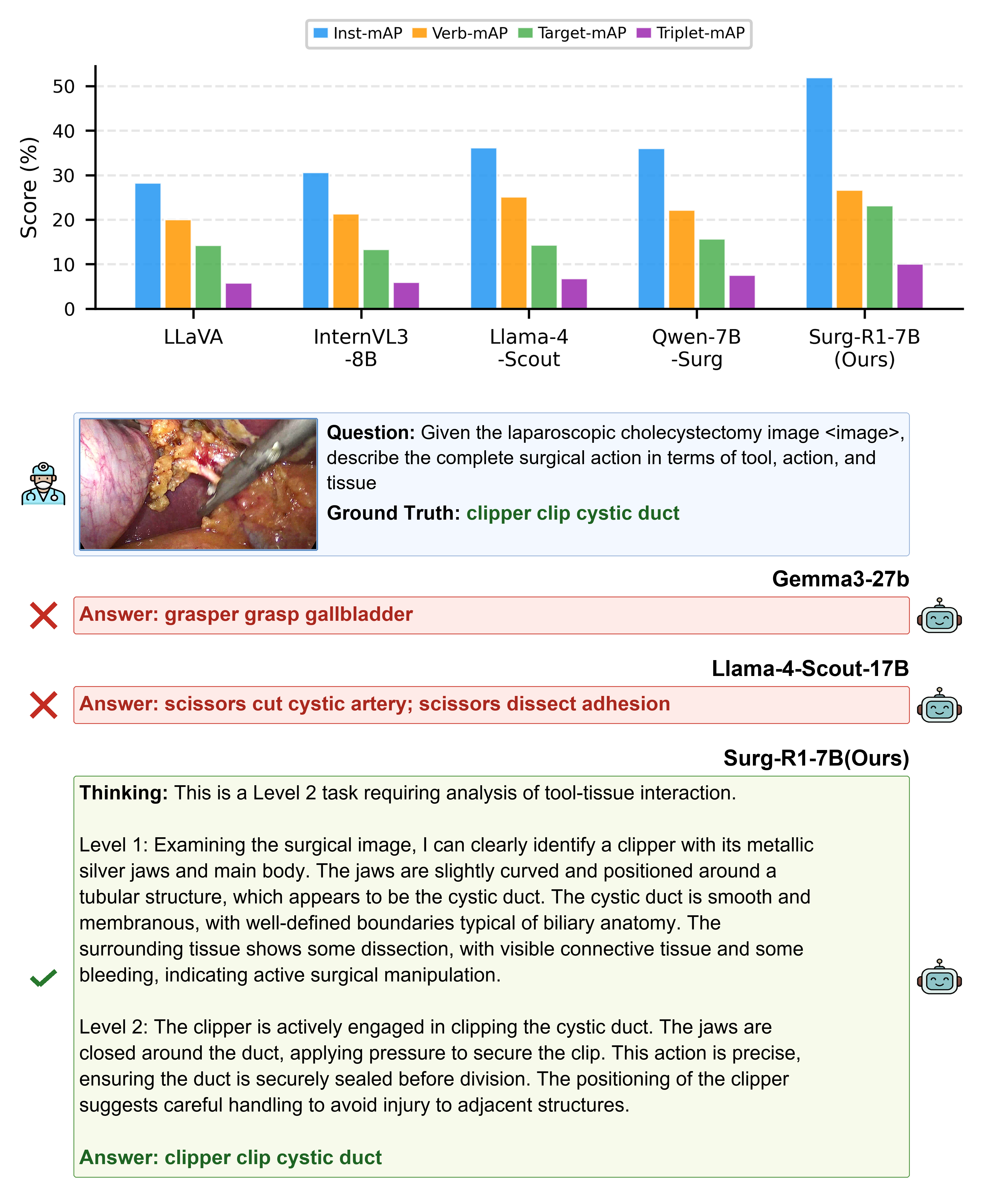}
\caption{{\bf Triplet recognition at Renji Hospital.} {\bf a.} Performance showing instrument, verb, target, and triplet mAP. {\bf b.} Example of triplet recognition for cystic duct clipping. Gemma3-27B predicts ``grasper grasp gallbladder,'' misidentifying both the instrument and target. Llama-4-Scout-17B predicts ``scissors cut cystic\_artery; scissors dissect adhesion,'' hallucinating multiple incorrect triplets. \modelFinal{} correctly identifies ``clipper clip cystic\_duct'' through hierarchical reasoning. Level 1 recognizes ``clipper with metallic silver jaws'' positioned around ``tubular structure'' identified as the cystic duct; Level 2 analyzes ``jaws closed around the duct, applying pressure to secure the clip.''}
\label{fig:renji_triplet_performance}
\end{figure}

On the Renji Hospital dataset (laparoscopic cholecystectomy) and the CUHK dataset (robotic surgery), \modelFinal{} demonstrates strong resilience against visual domain shifts in tool appearance and camera perspectives. For triplet recognition at Renji Hospital (Figure~\ref{fig:renji_triplet_performance}), general-purpose models severely degrade when encountering unfamiliar tissue textures and local lighting variations. As shown in the example, Gemma3-27B defaults to statistical priors (predicting ``grasper grasp gallbladder''), while Llama-4-Scout-17B hallucinates multiple non-existent actions (``scissors cut cystic artery; scissors dissect adhesion''). These failures stem from a breakdown in perceptual grounding, where models map ambiguous visual noise directly to high-frequency surgical triplets. \modelFinal{} avoids this structural collapse through strict hierarchical dependency. By first securing Level 1 morphological evidence (isolating the ``metallic silver jaws'' and a distinct ``tubular structure''), the model actively filters out the surrounding complex tissue noise. Consequently, Level 2 reasoning does not guess the interaction from the global scene context but specifically analyzes the grounded local kinematics (recognizing the ``jaws closed around the duct, applying pressure''). This ensures the extracted ``clipper clip cystic duct'' triplet is physically validated rather than statistically inferred. 

\begin{figure}[!htb]
\centering
\includegraphics[width=0.78\linewidth]{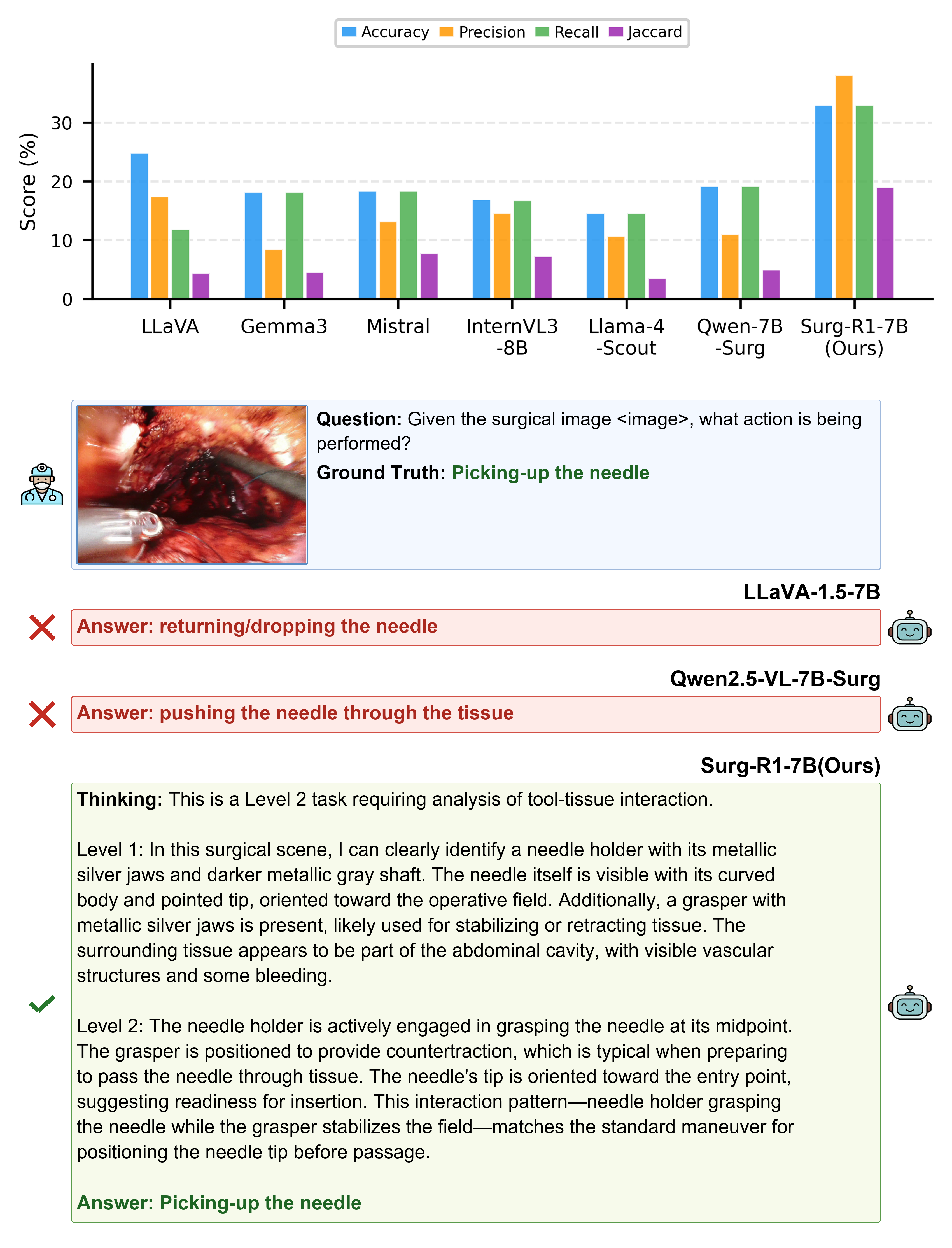}
\caption{{\bf Action recognition performance on CUHK external validation dataset.} {\bf a.} Performance comparison using accuracy, recall, precision, and F1 score. {\bf b.} Example of action recognition for needle manipulation. Baseline models confuse ``picking-up the needle'' with visually similar suturing gestures. \modelFinal{} correctly identifies the action through hierarchical reasoning. Level~1 recognizes tool types and spatial arrangement; Level~2 analyzes tool-tissue interaction patterns to distinguish the current action.}
\label{fig:cuhk_action_recognition}
\end{figure}

This invariant relational understanding is equally critical for fine-grained action recognition on the CUHK dataset (Figure~\ref{fig:cuhk_action_recognition}), where \modelFinal{} outperforms Qwen2.5-VL-7B-Surg by 13.8 percentage points by resolving severe spatial ambiguity. During the suturing phase of robotic-assisted radical prostatectomy, distinct kinematic actions (such as ``picking-up the needle'' versus active ``suturing'') frequently share nearly identical macroscopic visual layouts. Baseline models are frequently confounded by spatial proximity, erroneously predicting a suturing gesture simply because a needle driver is adjacent to tissue. \modelFinal{} overcomes this visual aliasing through its hierarchical architecture. Even when instrument tips are partially obscured or visually ambiguous, Level 1 perception successfully isolates the free-standing needle body and the secondary stabilizing tool. Subsequently, Level 2 reasoning investigates the overarching mechanical state. Rather than hallucinating an active tissue penetration based on mere spatial proximity, the model accurately infers the preparatory kinematics from the needle's orientation and the countertraction provided by the secondary grasper. By evaluating this broader interaction pattern associated with pre-suturing positioning, the model successfully identifies the ``picking-up'' action, demonstrating robust physical reasoning that compensates for localized visual uncertainties.

\subsubsection{Cross-Institutional Contextual Consistency.}
\begin{figure}[!htb]
\centering
\includegraphics[width=0.78\linewidth]{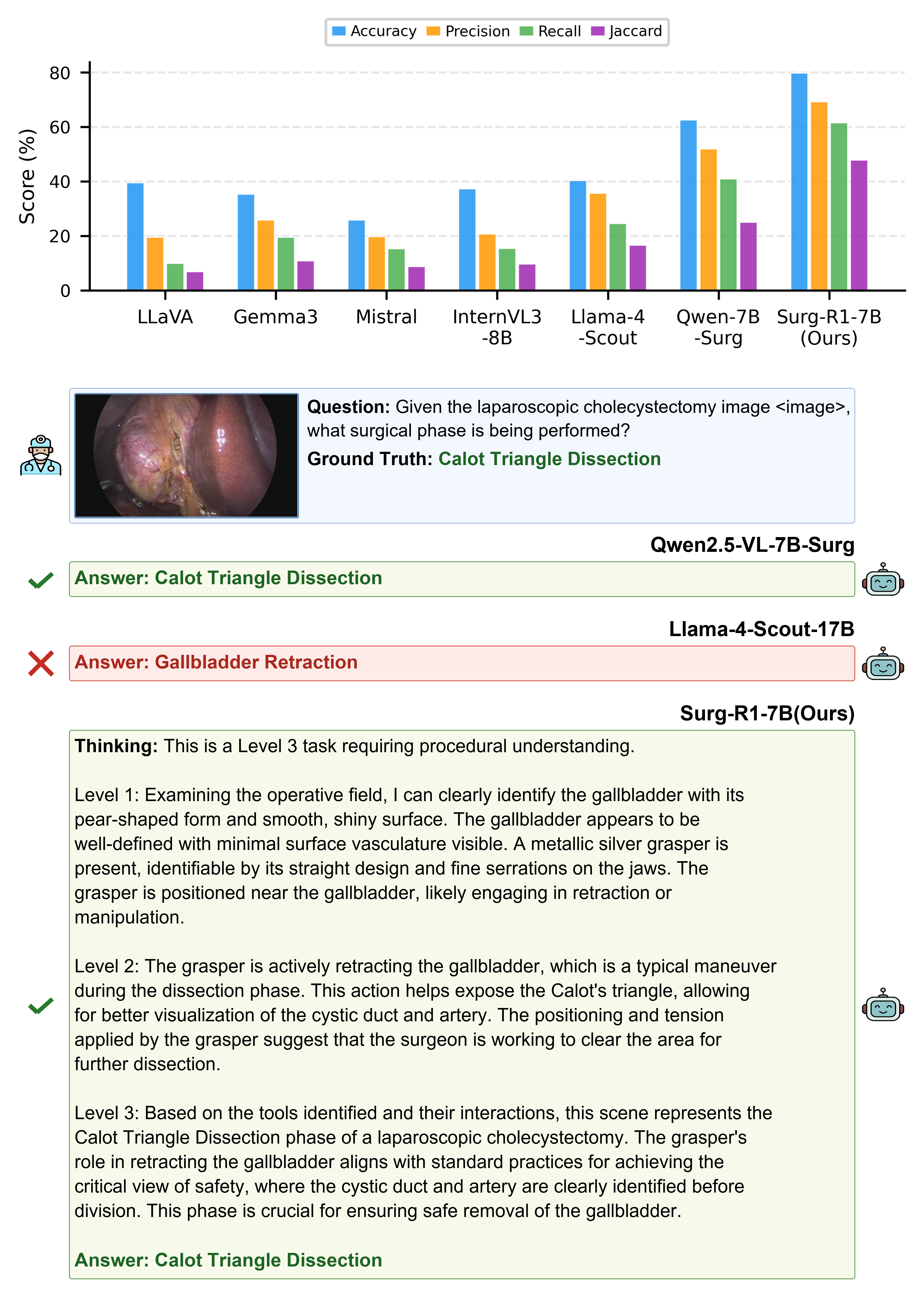}
\caption{{\bf Phase recognition on Strasbourg clinical dataset.} {\bf a.} Performance comparison using accuracy, precision, recall, and Jaccard index. {\bf b.} Example of phase recognition in European clinical setting. Qwen2.5-VL-7B-Surg correctly identifies ``Calot Triangle Dissection,'' while Llama-4-Scout-17B misclassifies it as ``Gallbladder Retraction.'' \modelFinal{} provides correct prediction with detailed reasoning. Level 1 identifies ``gallbladder with pear-shaped form'' and ``metallic silver grasper''; Level 2 recognizes ``grasper actively retracting the gallbladder to expose the Calot's triangle''; Level 3 concludes this represents the dissection phase where ``cystic duct and artery are clearly identified before division.''}
\label{fig:strasbourg_performance}
\end{figure}

\begin{figure}[!htb]
\centering
\includegraphics[width=0.78\linewidth]{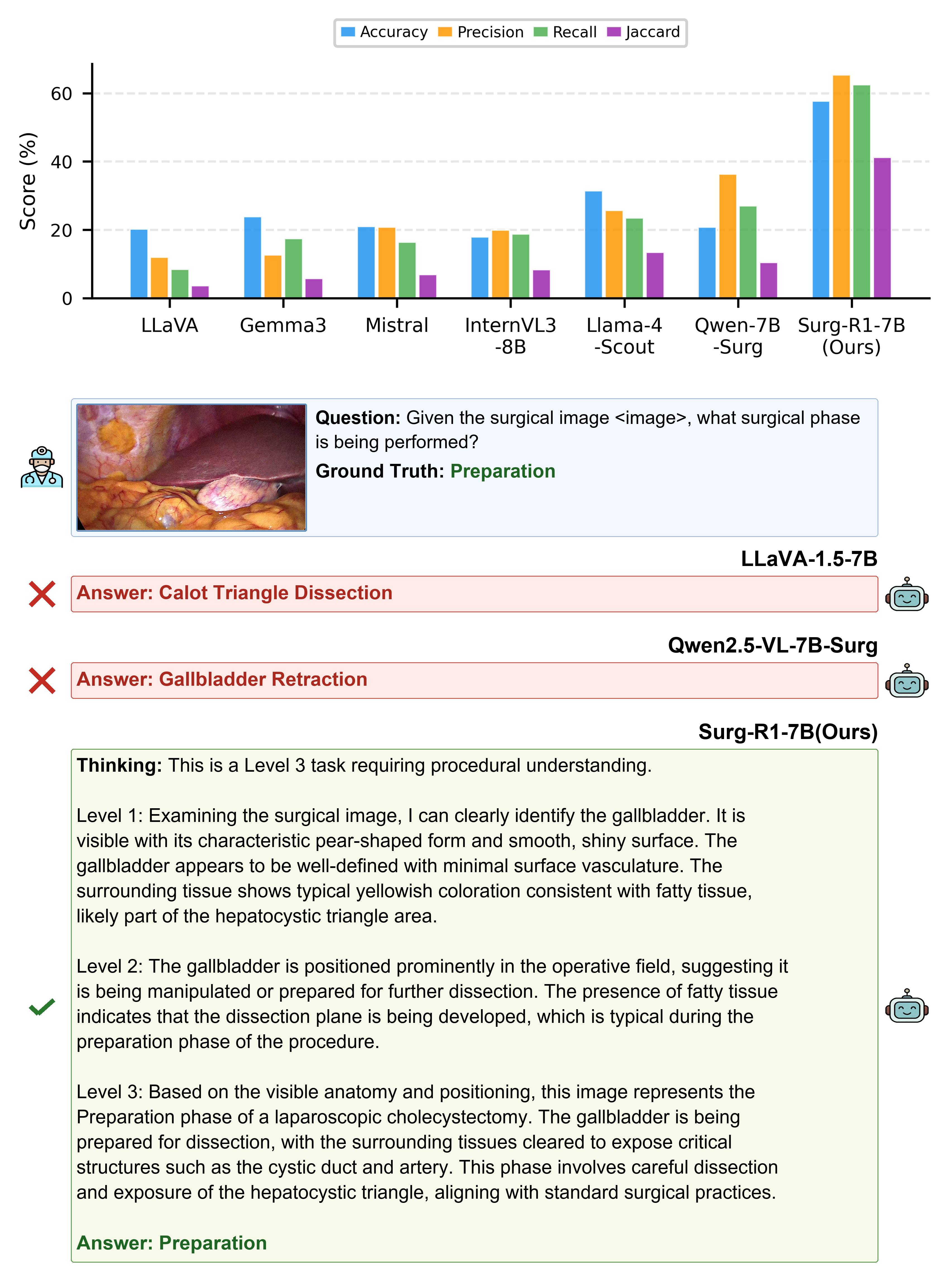}
\caption{{\bf Phase recognition at Renji Hospital.} {\bf a.} Performance comparison using accuracy, precision, recall, and Jaccard index. {\bf b.} Example of phase recognition in laparoscopic cholecystectomy. LLaVA-1.5-7B incorrectly predicts ``Calot Triangle Dissection,'' while Qwen2.5-VL-7B-Surg predicts ``Gallbladder Retraction,'' both misinterpreting the early procedural stage. \modelFinal{} correctly identifies ``Preparation'' phase through hierarchical reasoning. Level 1 recognizes the gallbladder with ``smooth, shiny surface'' and ``surrounding fatty tissue''; Level 2 notes the gallbladder ``positioned prominently'' being ``manipulated or prepared''; Level 3 concludes this represents the preparation phase where ``dissection plane is being developed.''}
\label{fig:renji_phase_performance}
\end{figure}

Phase recognition across different hospitals requires abstracting away from specific surgeon habits. On the Strasbourg dataset, \modelFinal{} achieves 79.80\% accuracy, a 17.2 percentage point improvement over the best baseline (Figure~\ref{fig:strasbourg_performance}). At Renji Hospital, it achieves a 36.9 percentage point improvement on an independently collected laparoscopic cholecystectomy dataset with distinct surgical techniques and imaging conditions (Figure~\ref{fig:renji_phase_performance}). In both settings, while baselines are confused by early-stage manipulations, \modelFinal{} utilizes Level 3 reasoning to correctly identify the ``Preparation'' phase by evaluating the contextual progression (e.g., recognizing that the dissection plane is still ``being developed''), proving that explicit cognitive structuring is superior to implicit pattern matching for cross-center clinical deployment.

\subsection{Comparison with Reasoning-Enhanced Models}
Table~\ref{tab:reasoning_comparison} compares \modelFinal{} against GPT-5.1 and Gemini 3.0 Pro, both of which employ advanced chain-of-thought capabilities. Compared to standard non-reasoning VLMs, these general-purpose reasoning models exhibit superior overall performance, validating the fundamental necessity of explicit reasoning for complex scene understanding. However, this general-purpose cognitive capability does not seamlessly transfer to specialized clinical domains. They achieve only 28.5\% and 29.8\% Arena Scores on surgical benchmarks, respectively. The performance gap is most pronounced on highly compositional tasks, such as triplet recognition, where GPT-5.1 achieves only 6.77\% compared with 51.69\% for \modelFinal{}. An in-depth analysis of the generated reasoning chains reveals two fundamental failure modes in general-purpose CoT when applied to surgical environments, both of which our hierarchical architecture explicitly resolves.

First, general reasoning models frequently suffer from eloquent hallucinations driven by spatial proximity. Because they lack domain-specific physical grounding, models like GPT-5.1 and Gemini 3.0 Pro consistently map visual adjacency directly to functional interaction. They erroneously predict active cutting, dissecting, or suturing simply because an instrument is located near an anatomical structure. \modelFinal{} overcomes this binding problem through its Level 2 relational reasoning, which strictly demands explicit kinematic and mechanical evidence (such as tissue tension, tool-tip orientation, or preparatory positioning) before asserting a surgical action. This enforces a physical validation step that prevents the model from generating plausible but clinically false narratives.

Second, general VLMs struggle to differentiate between isolated local actions and broader clinical objectives. As observed across phase recognition tasks, proprietary models tend to anchor primarily on localized tissue manipulation. They fail to recognize when a local action (such as simple retraction or stapling) is actively serving a higher-level procedural goal (such as transitioning into the dissection phase or creating a gastric pouch). By enforcing a transition to Level 3 contextual reasoning, \modelFinal{} successfully synthesizes local perceptual features and physical interactions into overarching surgical intent. These results indicate that unstructured, general-purpose reasoning capabilities are insufficient for high-stakes clinical scene understanding without a dedicated hierarchical cognitive framework.

\begin{table}[htbp]
\centering
\small
\caption{{\bf Performance comparison with reasoning-enhanced models.} Accuracy (\%) on representative tasks. Largest gap on triplet recognition (44.9 percentage points over GPT-5.1).}
\label{tab:reasoning_comparison}
\begin{tabular}{lccc}
\toprule
\textbf{Task} & \textbf{GPT-5.1} & \textbf{Gemini-3.0-Pro} & \textbf{Ours} \\
\midrule
CholecT50 Triplet & 6.77 & 8.20 & \textbf{51.69} \\
Cholec80 Phase & 49.70 & 52.35 & \textbf{80.90} \\
MultiBypass140 Phase & 29.01 & 28.72 & \textbf{49.94} \\
\bottomrule
\end{tabular}
\end{table}

%% file: 8_discussion.tex
We developed \modelFinal{}, a surgical VLM that integrates hierarchical CoT reasoning to understand complex surgical scenes. While general-purpose VLMs process visual and textual inputs effectively, they often struggle with the compositional nature of surgery, producing outputs without explaining the rationale. In addition, existing surgical VLMs are typically designed for single-task settings, limiting their ability to generalize across diverse surgical understanding tasks. \modelFinal{} addresses these limitations by introducing a structured, interpretable reasoning process, leveraging a four-stage pipeline with multi-task training. Evaluated across ten datasets, consisting of four public benchmarks and six external validation sets from five independent clinical centers, \modelFinal{} outperformed general reasoning models like GPT-5.1 and Gemini 3.0 Pro on the majority of evaluated tasks, as well as specialized surgical VLMs. It achieved an Arena Score of 57.7\% on public benchmarks compared to 29.8\% for Gemini 3.0 Pro and 28.5\% for GPT-5.1. This performance gap indicates that domain-specific reasoning structures provide benefits beyond what general-purpose models achieve.

\subsection{Unified Multi-Task Architecture}
It is important to contextualize \modelFinal{}  within the broader landscape of task-specific surgical AI methods. Community challenges have driven substantial progress on individual tasks, such as triplet recognition in the CholecTriplet challenges \cite{nwoye2023cholectriplet2021,nwoye2023cholectriplet2022} and CVS assessment in the SAGES CVS Challenge \cite{alapatt2025sages_cvs_challenge}, with top-performing methods achieving strong performance through architectures optimized for a single objective (e.g., using temporal modeling over videos or multi-model ensembles). However, \modelFinal{}  represents a fundamentally different paradigm. Rather than relying on narrow, specialized pipelines, it operates as a unified multi-task VLM that addresses triplet recognition, CVS assessment, phase recognition, action recognition, and instrument localization simultaneously on individual frames with explicit text-based reasoning. While direct numerical comparison is complex due to different evaluation protocols and input modalities, the primary advantage of leveraging a VLM for surgical analysis lies in its cross-task generalizability. Task-specific pipelines are optimized for leveraging temporal context and specialized representations, whereas our proposed unified VLM, \modelFinal{}, provides interpretable reasoning chains and the unique ability to handle multiple, diverse clinical tasks simultaneously without requiring architectural redesign.

\subsection{Interpretability for Clinical Trust and Hallucination Mitigation}
Another significant advantage of \modelFinal{} is its ability to reason in a manner consistent with surgical decision-making. For AI to be safely integrated into the operating room, accuracy must be paired with trustworthiness. Black-box models that generate predictions without traceable logic face significant barriers to clinical adoption. This issue is particularly evident in tasks like triplet recognition, a compositional task benchmarked through community challenges~\cite{nwoye2023cholectriplet2021,nwoye2023cholectriplet2022} that require recognizing $\langle$instrument, verb, target$\rangle$ tuples, where general models frequently hallucinate interactions based solely on visual co-occurrence. \modelFinal{} addresses this by enforcing a strict three-level reasoning hierarchy: \textbf{perceptual grounding}, \textbf{relational understanding}, and \textbf{contextual reasoning}. By requiring the model to first explicitly identify tools and anatomical structures by their physical properties (e.g., "metallic jaws") before analyzing mechanical interactions, it produces logical and verifiable rationales. This transparency not only improves accuracy (51.69\% vs. 6.77\% for GPT-5.1 on triplet recognition) but also allows surgeons to dynamically validate the logic against their own expertise, a critical requirement for clinical trust.

\subsection{Robust Generalization across Heterogeneous Clinical Environments}
Crucially, developing high-level structured reasoning poses two major scalability challenges: the reliance on dense process supervision during training and the prohibitive cost of generating manual annotations on reasoning data. To address the training constraint, the model employs GRPO for reinforcement learning. By using only final outcome labels as reward signals, GRPO refines the reasoning capabilities of the model without the need for exhaustive, step-by-step process supervision. Furthermore, \modelFinal{} overcomes the data bottleneck through an iterative refinement pipeline. This approach leverages rejection sampling and knowledge distillation to automatically scale an initial seed of 42,000 GPT-generated samples into approximately 320,000 high-quality CoT pairs. This massive, automated expansion of reasoning data provides \modelFinal{} with highly robust generalization capabilities across diverse clinical environments. Deep learning models often experience severe performance drops when faced with out-of-distribution (OOD) data, such as variations in endoscopic lighting, camera angles, or institution-specific surgical techniques. However, in our multi-center evaluation, \modelFinal{} maintained a 15.2\% improvement over domain-adapted baselines on external data. Notably, it achieved a 36.9\% gain at Renji Hospital, where institutional differences in surgical technique, equipment, and imaging conditions represent real-world distribution shifts. This suggests that instead of merely memorizing dataset-specific pixel patterns, \modelFinal{} has learned transferable surgical concepts. By reasoning through underlying anatomical principles, such as identifying continuous tissue connections despite unfamiliar shadows, the model proves its reliability for deployment across heterogeneous real-world hospital networks.

\subsection{Potential Clinical Impacts}
The translational potential of \modelFinal{} extends beyond automated video analysis; it introduces new possibilities across three key areas in the clinic:
(1) \textbf{Surgical Report Generation}: Leveraging the language model's deep understanding of the surgical scene and procedural workflow, \modelFinal{}  has the potential to automate the tedious documentation process. By continuously analyzing the procedure, it can be developed to automatically draft comprehensive operative notes and surgical reports in real-time, reducing the administrative burden on surgical teams. (2) \textbf{Surgical Education}: The transparent reasoning provided by \modelFinal{}  can function as an active educational tool in the operating room. Rather than simply outputting a prediction, the model verbalizes the anatomical and procedural cues that define a specific action or phase transition. This explicit guidance can accelerate the learning process for junior surgeons, helping to narrow the experience gap between novices and experts. (3) \textbf{Intraoperative Safety}: As a real-time safety monitor, the high accuracy of \modelFinal{}  in assessing the Critical View of Safety approaches expert levels, recording 92.50\% overall accuracy at West China Hospital and 87.36\% at Nanfang Hospital. The model can help prevent severe complications, such as bile duct injuries, by offering objective, safety-critical verification during the procedure without requiring secondary, task-specific detection pipelines.

\subsection{Limitations and Future Directions}
Several limitations warrant consideration. First, the iterative training pipeline demands substantial computational resources, which may limit adoption in resource-constrained settings. Second, formal clinical validation of the reasoning outputs by surgeons is necessary before deployment, as model-generated rationales must meet the interpretability and reliability standards required for intraoperative decision support. Third, the hierarchical CoT reasoning introduces additional inference latency compared with direct-prediction models, and further optimization is needed to meet real-time requirements in the operating room. Looking ahead, the demonstrated cross-institutional robustness across laparoscopic cholecystectomy, robotic prostatectomy, and gastric bypass motivates extension to fundamentally different surgical paradigms, including open surgery, microsurgery, and non-abdominal specialties, where visual characteristics and instrument appearances differ substantially from the current training domain. Future work should also address integration with robotic control systems, prospective clinical studies evaluating the impact of reasoning-based decision support on surgical outcomes, and application to additional medical imaging domains such as pathology and radiology. Structured domain-specific reasoning combined with reinforcement learning represents a viable path toward interpretable and generalizable surgical AI.

%% file: 5_method.tex
\indent We train \textbf{\modelFinal} through a four-stage pipeline. \textbf{Stage 1} establishes vision-language alignment via supervised fine-tuning (SFT); \textbf{Stage 2} introduces hierarchical chain-of-thought (CoT) reasoning through cold-start SFT; \textbf{Stage 3} refines reasoning using GRPO-based reinforcement learning; and \textbf{Stage 4} enables iterative refinement through rejection sampling and knowledge distillation. All stages use LoRA-based parameter-efficient fine-tuning for both the vision encoder and LLM decoder.

\subsection{Hierarchical Chain-of-Thought Design}

\indent We design a three-level hierarchical CoT structure. \textbf{Level 1 (Perceptual Grounding)} identifies and localizes visual elements: tool recognition, tissue recognition, tool localization, tissue localization, grid position recognition, and CVS criteria assessment. \textbf{Level 2 (Relational Understanding)} analyzes behaviors and relationships among tools, actions, and tissues: action recognition, motion recognition, and triplet extraction. \textbf{Level 3 (Contextual Reasoning)} uses lower-level information for image-level context: tool counting, step recognition, phase recognition, and overall CVS assessment. The complete CoT output follows the format: \texttt{<think>} Level 1 reasoning $\rightarrow$ Level 2 reasoning $\rightarrow$ Level 3 reasoning \texttt{</think>} \texttt{<answer>} Final Answer \texttt{</answer>}.

\subsection{Stage 1: Base SFT Initialization}

\indent The first stage establishes surgical vision-language alignment using VQA-format labels (e.g., ``Current surgical phase is Calot Triangle Dissection''). We use Qwen2.5-VL-7B as the base model and construct $\mathcal{D}_{\text{Base}}$ (\datasetBase{}) from multiple public surgical benchmarks, covering tool/tissue/action/phase recognition, tool localization, and CVS assessment.

Given a surgical image $I$ and textual instruction $T$, the visual encoder $\mathcal{E}_v$ extracts visual embeddings $Z_v = \mathcal{E}_v(I)$, which are projected into the language space via an MLP-based merger $\mathcal{M}$: $Z_v' = \mathcal{M}(Z_v)$. The fused representation $Z = [Z_v'; Z_t]$ is processed by the transformer decoder to generate the output. Training minimizes the standard cross-entropy loss:
\begin{equation}
\mathcal{L}_{\text{CE}} = - \sum_{i=1}^{L_o} \log p(y_i \mid y_{<i}, I, T).
\end{equation}
This stage provides foundational surgical perception capabilities without explicit reasoning.

\subsection{Stage 2: Hierarchical CoT Cold-Start SFT}

\indent The second stage embeds structured reasoning priors into the model using the 3-Level CoT labels. We construct $\mathcal{D}_{\text{Cold\_Start}}$, the GPT-generated cold-start subset of \datasetCoT{}, through an automated pipeline that queries GPT-5.1 with carefully designed \textit{surgery-wise generation constraints}.

\paragraph{Surgery-Aware CoT Synthesis.}
We provide GPT-5.1 with structured domain knowledge for each procedure type: tool and tissue lists with color, shape, and texture characteristics; action and phase vocabularies; and co-occurrence statistics describing typical tool--tissue--action patterns. The prompt instructs the model to describe what is observed in the image, and when visual observations conflict with reference descriptions, the model trusts visual evidence. Although GPT-5.1 receives the ground-truth label as an implicit target, it is instructed to derive the answer through visual observation and hierarchical reasoning rather than reverse-engineering from the answer. This forward-reasoning approach ensures the generated CoT reflects genuine visual grounding.

This pipeline yields approximately 42,000 samples covering six tasks: Cholec80 phase recognition, CholecT50 triplet recognition, EndoScape2023 CVS detection, RARP action recognition, Multibypass phase recognition, and Sleeve triplet recognition. Cold-start SFT on $\mathcal{D}_{\text{Cold\_Start}}$ establishes preliminary hierarchical reasoning capabilities and provides the foundation for subsequent RL refinement.

\subsection{Stage 3: GRPO-Based RL Refinement}

\indent The second stage uses GPT-5.1-generated CoT labels, which are costly to produce at scale. The third stage employs Group Relative Policy Optimization (GRPO), a critic-free reinforcement learning algorithm that requires only \textbf{raw labels} (e.g., ``Calot Triangle Dissection'') as reward signals. This enables training on $\mathcal{D}_{\text{GRPO}}$ without generating CoT annotations for all samples.

\paragraph{GRPO Objective.}
For each input $x$, GRPO samples a group of $G$ candidate outputs $\{o_1, o_2, \ldots, o_G\}$ from the current policy $\pi_\theta$. The policy is updated by maximizing the following objective:
\begin{equation}
\mathcal{J}_{\text{GRPO}}(\theta) = \mathbb{E}_{x \sim \mathcal{D}, \{o_i\}_{i=1}^G \sim \pi_\theta(\cdot|x)} \left[ \frac{1}{G} \sum_{i=1}^{G} \hat{A}_i \cdot \log \pi_\theta(o_i | x) \right],
\end{equation}
where the group-normalized advantage $\hat{A}_i$ is computed as:
\begin{equation}
\hat{A}_i = \frac{r(o_i) - \mu_r}{\sigma_r + \epsilon}, 
\mu_r = \frac{1}{G}\sum_{j=1}^{G} r(o_j), \quad \sigma_r = \sqrt{\frac{1}{G}\sum_{j=1}^{G} (r(o_j) - \mu_r)^2}.
\end{equation}
This formulation eliminates the need for an explicit critic network by using group statistics for baseline estimation.

\paragraph{Reward Design.}
We adopt a lightweight reward function with three components:
\begin{equation}
r(o) = \lambda_f \cdot \mathbb{I}_{\text{format}}(o) + \lambda_a \cdot \mathbb{I}_{\text{answer}}(o) + \lambda_s \cdot \mathbb{I}_{\text{structure}}(o),
\end{equation}
where $\mathbb{I}_{\text{format}}(o)$ indicates whether the output follows the correct hierarchical CoT format with proper \texttt{<think>} and \texttt{<answer>} tags, $\mathbb{I}_{\text{answer}}(o)$ indicates exact match between the extracted final answer and the ground-truth raw label, and $\mathbb{I}_{\text{structure}}(o)$ indicates whether ``Level 1'' appears in the reasoning trace to ensure perceptual grounding precedes higher-level reasoning. The structure reward ($\lambda_s > 0$) is applied only during initial training to establish reasoning patterns, then disabled ($\lambda_s = 0$) to focus on answer correctness.

\paragraph{High-Entropy Token Focusing.}
High-entropy tokens shape reasoning trajectories; we bias GRPO updates toward uncertain positions. For each token position $t$, we compute the entropy of the output distribution:
\begin{equation}
H_t = -\sum_{v \in \mathcal{V}} p_\theta(v | o_{<t}, x) \log p_\theta(v | o_{<t}, x).
\end{equation}
We then apply an entropy-based weight $w_t$ to concentrate learning signals on high-entropy positions:
\begin{equation}
w_t = \frac{\exp(H_t / \tau)}{\frac{1}{L}\sum_{j=1}^{L} \exp(H_j / \tau)},
\end{equation}
where $\tau$ is a temperature hyperparameter and $L$ is the sequence length. The weighted GRPO objective becomes:
\begin{equation}
\mathcal{J}_{\text{GRPO}}^{\text{HE}}(\theta) = \mathbb{E} \left[ \frac{1}{G} \sum_{i=1}^{G} \hat{A}_i \cdot \sum_{t=1}^{L} w_t \cdot \log \pi_\theta(o_{i,t} | o_{i,<t}, x) \right].
\end{equation}
This entropy-weighted update focuses optimization on decision-critical tokens.

\subsection{Stage 4: Iterative Refinement}

\indent The fourth stage implements iterative refinement through a dual-pathway mechanism: \textit{rejection sampling} generates pseudo-labels for samples the model predicts correctly, while \textit{knowledge distillation} from a teacher model handles challenging samples. This expands reasoning capabilities beyond the initial CoT dataset. Each iteration proceeds as follows.

\paragraph{Step 1: Hard Sample Identification.}
For each training sample $x$ with ground-truth label $y$, we perform $K=3$ independent rollouts using the current model and determine whether it is a \textit{hard sample} based on prediction consistency:
\begin{equation}
\text{HardSample}(x) = \mathbb{I}\left[\sum_{k=1}^{K} \mathbb{I}[\hat{y}_k = y] = 0\right],
\end{equation}
where $\hat{y}_k$ denotes the predicted answer from the $k$-th rollout. A sample is classified as hard if all three predictions fail to match the ground truth.

\paragraph{Step 2: Pseudo-Label Generation via Dual Pathways.}
For non-hard samples, the latest fine-tuned VLM performs rollout to generate 3-Level CoT predictions; if the final answer matches the ground-truth label, the reasoning trace is accepted as a \textit{Predicted Pseudo Label}, otherwise the sample is re-tagged as hard. For hard samples, a commercial VLM is queried with the same surgery-wise generation constraints used in Stage~2 to produce a \textit{Generated Pseudo Label}.

\paragraph{Step 3: Distillation.}
The collected pseudo-labels from both pathways are aggregated to fine-tune the next iteration of the VLM. This process repeats for $N$ iterations, progressively improving performance on challenging samples while maintaining quality on easier ones. The complete CoT training data produced across Stages~2--4, comprising $\mathcal{D}_{\text{Cold\_Start}}$ and all iteratively refined pseudo-labels, collectively constitutes \datasetCoT{}.

%% file: 9_supp.tex

This supplementary document provides additional details on the hierarchical reasoning design, training pipeline, datasets, training data statistics, and comprehensive experimental results across all benchmarks.

\subsection{Three-Level Reasoning Hierarchy}

\begin{figure}[htbp]
\centering
\includegraphics[width=0.95\linewidth]{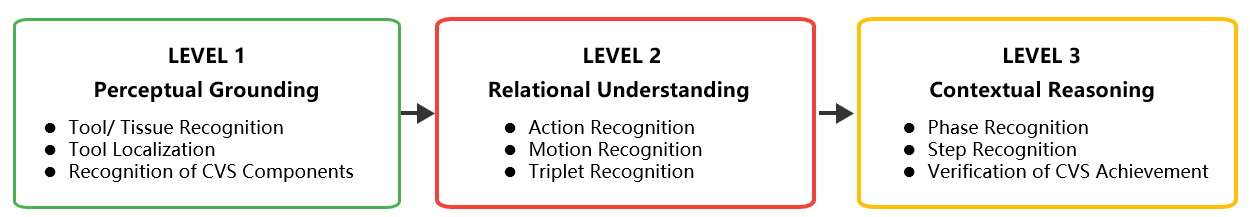}
\caption{{\bf Three-level reasoning hierarchy.} Level 1: visual element identification (tools, tissues, spatial relationships). Level 2: tool-tissue-action interaction analysis and triplet recognition. Level 3: procedural workflow assessment including phase recognition, step recognition, and CVS assessment.}
\label{fig:cot_hierarchy}
\end{figure}

\subsection{Training Pipeline}

\begin{figure}[htbp]
\centering
\includegraphics[width=0.95\linewidth]{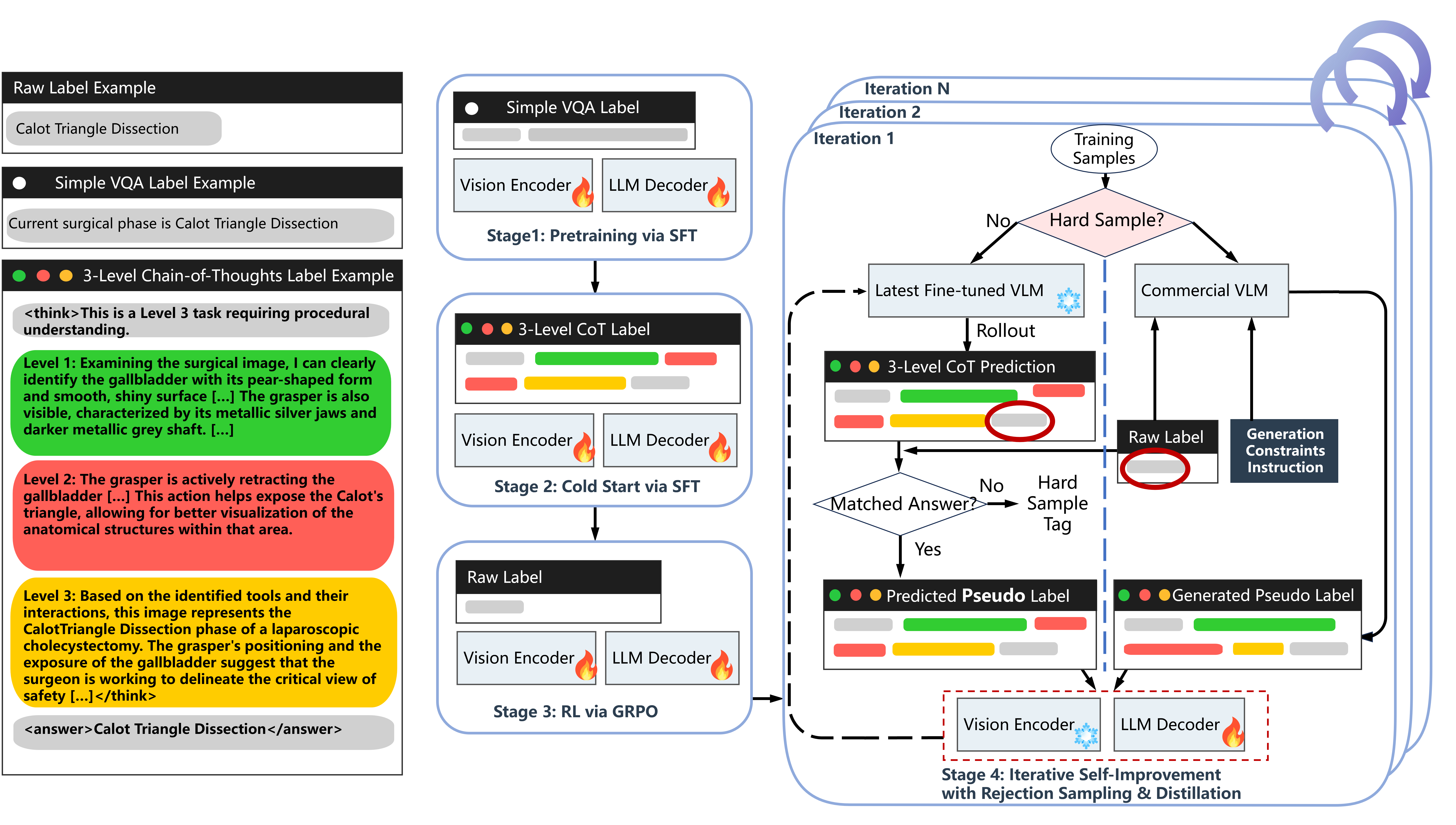}
\caption{{\bf Four-stage training pipeline for \modelFinal{}.} Stage 1: base supervised fine-tuning establishes vision-language alignment on surgical VQA data. Stage 2: chain-of-thought cold-start training introduces hierarchical reasoning structure using surgery-aware CoT synthesis. Stage 3: GRPO-based reinforcement learning refines reasoning robustness using only outcome labels. Stage 4: iterative refinement expands reasoning coverage through rejection sampling and knowledge distillation.}
\label{fig:pipeline_supp}
\end{figure}

The four-stage pipeline progressively builds surgical reasoning capabilities: vision-language alignment (Stage 1), structured reasoning introduction (Stage 2), reasoning robustness (Stage 3), and coverage expansion (Stage 4). Each stage builds on preceding capabilities, with LoRA-based parameter-efficient fine-tuning applied throughout.

\subsection{Dataset Information}

\subsubsection{Pre-training Data.}

The pre-training dataset \datasetBase{} aggregates 23 public surgical datasets spanning six task types: tool localization, surgical VQA, triplet recognition, CVS assessment, phase recognition, and action recognition. The dataset composition is visualized in Figure~\ref{fig:arena_scores}a, with image counts ranging from 319 (SegCol-Endovis) to over 2 million (MultiBypass140 for phase recognition). This diverse collection ensures broad anatomical coverage and task variety for establishing foundational surgical perception capabilities.

\subsubsection{Public Benchmark Datasets.}

\begin{table*}[h]
    \centering
    \small
    \caption{Public Dataset Surgery Type and Download Links}
    \label{tab:dataset_links}
    \begin{tabular}{l l p{7.5cm}} 
        \toprule
        \textbf{Dataset Name} & \textbf{Surgery Type} & \textbf{Download URL} \\
        \midrule
        \SegColCite & Colonoscopy & \url{https://www.synapse.org/Synapse:syn54124209} \\
        \CholecEightyCite\ (Phase+Tool) & Laparoscopic Cholecystectomy & \url{https://camma.unistra.fr/datasets/} \\
        \CholecEightyCVSCite & Laparoscopic Cholecystectomy & \url{https://camma.unistra.fr/datasets/} \\
        \CholecFiftyCite & Laparoscopic Cholecystectomy & \url{https://github.com/CAMMA-public/cholect50} \\
        \AutoLaparoCite\ (Phase) & Laparoscopic Hysterectomy & \url{https://github.com/ziyiwangx/AutoLaparo} \\
        \AutoLaparoCite\ (Seg) & Laparoscopic Hysterectomy & \url{https://github.com/ziyiwangx/AutoLaparo} \\
        \AutoLaparoCite\ (Motion) & Laparoscopic Hysterectomy & \url{https://github.com/ziyiwangx/AutoLaparo} \\
        \CholecEightyVQACite & Laparoscopic Cholecystectomy & \url{https://github.com/lalithjets/SurgicalGPT} \\
        \PSIAVAVQACite & Robot-Assisted Radical Prostatectomy & \url{https://github.com/lalithjets/SurgicalGPT} \\
        \EndoVisEighteenVQACite & Robot-Assisted Nephrectomy & \url{https://github.com/lalithjets/SurgicalGPT} \\
        \CoPESDCite & Endoscopic Submucosal Dissection (ESD) & \url{https://github.com/gkw0010/CoPESD} \\
        \DSADCite\ (Multi) & Laparoscopic Abdominal Surgery & \url{https://www.kaggle.com/datasets/anindyamajumder/the-dresden-surgical-anatomy-dataset} \\
        \DSADCite\ (Single) & Laparoscopic Abdominal Surgery & \url{https://www.kaggle.com/datasets/anindyamajumder/the-dresden-surgical-anatomy-dataset} \\
        \EndoscapesCite\ (Seg) & Laparoscopic Cholecystectomy & \url{https://github.com/CAMMA-public/Endoscapes} \\
        \GraSPCite\ (LongTerm) & Robot-Assisted Radical Prostatectomy & \url{https://github.com/BCV-Uniandes/GraSP} \\
        \MultiBypassCite & Laparoscopic Gastric Bypass & \url{https://github.com/CAMMA-public/MultiBypass140} \\
        \PitVQACite & Transsphenoidal Pituitary Surgery & \url{https://github.com/mobarakol/PitVQA} \\
        \CholecInstanceSegCite & Laparoscopic Cholecystectomy & \url{https://github.com/labdeeman7/cholec_instance_seg} \\
        \CholecTrackCite & Laparoscopic Cholecystectomy & \url{https://github.com/CAMMA-public/cholectrack20} \\
        \SARRARPCite\ (Seg) & Robot-Assisted Radical Prostatectomy & \url{https://www.synapse.org/Synapse:syn27618412} \\
        \SARRARPCite\ (Action) & Robot-Assisted Radical Prostatectomy & \url{https://www.synapse.org/Synapse:syn27618412} \\
        \bottomrule
    \end{tabular}
\end{table*}

\begin{table}[h]
\centering
\caption{Statistics of conversations generated by GPT and \modelName{} across different datasets.}
\label{tab:dataset_stats}
\begin{tabular}{lrrr}
\toprule
\textbf{Dataset} & \textbf{GPT Generated} & \textbf{\modelName{} Generated} & \textbf{Total Conversations} \\
\midrule
\CholecEightyCite{} + \CholecFiftyCite & 9,262 & 103,492 & 112,754 \\
\MultiBypassCite & 6,000 & 71,479 & 77,479 \\
\SARRARPCite & 5,975 & 64,530 & 70,505 \\
\EndoscapesCite & 6,000 & 37,857 & 43,857 \\
\EndoVisEighteenVQACite & 11,862 & 2,244 & 14,106 \\
\EndoVisSeventeenCite & 1,776 & 0 & 1,776 \\
\midrule
\textbf{Total} & \textbf{40,875} & \textbf{279,602} & \textbf{320,477} \\
\bottomrule
\end{tabular}
\end{table}

\subsection{Training Data Statistics}

Table~\ref{tab:dataset_stats} summarizes the chain-of-thought conversation data generated for training \modelFinal{}. The dataset comprises approximately 320,000 conversations, with roughly 41,000 high-quality samples generated using GPT-5.1 following our surgery-aware CoT synthesis pipeline, and the remainder generated through iterative refinement (combining model-generated and teacher-distilled samples).

\subsection{Arena Score Calculation}

The Arena Score provides a unified performance metric for comparing models across diverse surgical AI tasks. For each model, we compute the Arena Score as the arithmetic mean of its performance scores across all available tasks:

\begin{equation}
\text{Arena Score} = \frac{1}{|T_m|} \sum_{t \in T_m} s_{m,t}
\end{equation}

\noindent where $T_m$ denotes the set of tasks for which model $m$ has valid results, and $s_{m,t}$ represents the primary performance metric for model $m$ on task $t$. The primary metric varies by task type:

\begin{itemize}
\item \textbf{Triplet Recognition} (CholecT50, Renji): Triplet Accuracy (\%)
\item \textbf{Phase Recognition} (Cholec80, MultiBypass140, Renji, Strasbourg): Accuracy (\%)
\item \textbf{Action Recognition} (SAR-RARP50, CUHK): Accuracy (\%)
\item \textbf{CVS Assessment} (West China, SMU): Overall Accuracy (\%)
\end{itemize}

\noindent We note that conventional CVS classifiers typically report mean Average Precision (mAP), which requires continuous probability outputs. Because our VLM generates categorical text responses (``achieved'' or ``not achieved'') for each criterion rather than continuous probability scores, probability-based metrics such as mAP are not applicable. We therefore report accuracy, which directly reflects the clinical decision scenario of binary criterion assessment.

This averaging approach provides a balanced comparison across tasks while naturally handling missing evaluations (e.g., proprietary models that could not be evaluated on certain benchmarks due to API limitations). Models are ranked by their Arena Score within each benchmark category (public vs. external validation).

\subsection{Detailed Experimental Results on Public Benchmarks}

This section presents comprehensive performance comparisons across all public benchmark datasets. Tables include results for general-purpose VLMs, proprietary reasoning models, and specialized surgical VLMs.

\begin{table*}[h!]
\centering
\caption{{\bf Triplet recognition performance on the CholecT50 dataset.} CholecT50 comprises 50 laparoscopic cholecystectomy (gallbladder removal) procedures annotated with instrument-verb-target triplets. Accuracy and mAP are reported in percentage (\%). Best results are \textbf{bolded}.}
\begin{tabular}{llcccccccc}
\toprule
 & & \multicolumn{8}{c}{CholecT50 Triplet Recognition} \\
\cmidrule(lr){3-10}
 & & \multicolumn{4}{c}{Accuracy (\%)} & \multicolumn{4}{c}{mAP (\%)} \\
\cmidrule(lr){3-6} \cmidrule(lr){7-10}
Model & Thinking & Inst. & Verb & Target & Triplet & Inst. & Verb & Target & Triplet \\
\midrule
\ModelLLaVAcite{} & -- & 3.69 & 9.41 & 0.23 & 0.00 & 22.48 & 14.12 & 9.41 & 2.35 \\
\ModelPhicite{} & -- & 14.83 & 12.98 & 3.46 & 0.12 & 22.32 & 13.99 & 9.39 & 2.35 \\
\ModelMistralFullCite{} & -- & 9.81 & 8.02 & 5.48 & 0.46 & 22.43 & 14.05 & 9.41 & 2.35 \\
\ModelInternVLSmallcite{} & -- & 51.24 & 8.94 & 26.20 & 2.08 & 22.35 & 14.01 & 9.70 & 2.39 \\
\ModelInternVLLargeCite{} & -- & 38.14 & 9.52 & 3.29 & 0.52 & 22.74 & 14.08 & 9.44 & 2.36 \\
\ModelMiniCPMVcite{} & -- & 18.70 & 13.04 & 1.62 & 0.00 & 22.54 & 14.20 & 9.42 & 2.35 \\
\ModelMiniCPMOcite{} & -- & 24.99 & 7.79 & 5.71 & 0.06 & 22.49 & 14.02 & 9.40 & 2.35 \\
\ModelGemmacite{} & -- & 8.89 & 7.39 & 1.56 & 0.06 & 22.44 & 14.01 & 9.44 & 2.35 \\
\ModelSkyworkCite{} & -- & 19.85 & 9.69 & 1.67 & 0.00 & 22.42 & 14.02 & 9.41 & 2.35 \\
\ModelLlamaFullCite{} & -- & 9.46 & 6.17 & 9.64 & 0.58 & 22.73 & 14.22 & 9.64 & 2.37 \\
\ModelQwenSmallcite{} & -- & 10.21 & 4.90 & 5.54 & 0.35 & 22.67 & 14.14 & 9.47 & 2.36 \\
\ModelQwenMediumcite{} & -- & 27.81 & 7.91 & 3.12 & 0.98 & 22.79 & 14.68 & 10.09 & 2.53 \\
\ModelQwenLargecite{} & -- & 32.66 & 7.91 & 5.25 & 1.27 & 23.38 & 14.45 & 10.86 & 2.71 \\
\ModelGPTFourOFullOCite{} & -- & 13.33 & 5.89 & 5.94 & 1.50 & 22.80 & 14.34 & 10.42 & 2.60 \\
\ModelQwenMaxcite{} & -- & 7.21 & 4.85 & 5.94 & 0.35 & 22.40 & 14.12 & 9.59 & 2.39 \\
\ModelGeminiFlashCite{} & -- & 15.18 & 7.44 & 15.87 & 1.85 & 23.14 & 14.45 & 10.91 & 2.54 \\
\midrule
\ModelGeminiProThinkCite{} & \checkmark & 24.32 & 52.86 & 12.76 &8.2& 39.29& 28.99 & 16.29& 6.86 \\
\ModelGPTFiveThinkCite{} & \checkmark & 49.31 & 45.61 & 12.91 & 6.77 & 31.73 & 22.65 & 12.54 & 4.59 \\
\midrule
\ModelQwenSurgCite{} & -- & 54.75 & 43.67 & 11.08 & 8.01 & 42.39 & 28.96 & 15.15 & 5.69 \\
\midrule
\textbf{Surg-R1 (Ours)} & \checkmark & \textbf{86.50} & \textbf{72.15} & \textbf{57.49} & \textbf{51.69} & \textbf{87.77} & \textbf{57.16} & \textbf{29.40} & \textbf{16.17} \\
\bottomrule
\end{tabular}
\label{tab:cholect50_triplet_supp}
\end{table*}

\begin{table*}[h!]
\centering
\caption{{\bf Surgical phase recognition performance on the Cholec80 dataset.} Cholec80 contains 80 laparoscopic cholecystectomy procedures with seven annotated surgical phases. All metrics are reported in percentage (\%). Best results are \textbf{bolded}.}
\begin{tabular}{llcccc}
\toprule
 & & \multicolumn{4}{c}{Cholec80 Phase Recognition (\%)} \\
\cmidrule(lr){3-6}
Model & Thinking & Accuracy & Recall & Precision & Jaccard \\
\midrule
\ModelLLaVAcite{} & -- & 23.46 & 14.22 & 11.53 & 6.31 \\
\ModelPhicite{} & -- & 22.45 & 14.83 & 12.16 & 5.55 \\
\ModelMistralFullCite{} & -- & 22.61 & 19.87 & 26.55 & 9.92 \\
\ModelInternVLSmallcite{} & -- & 23.88 & 15.21 & 15.19 & 6.67 \\
\ModelInternVLLargeCite{} & -- & 27.32 & 24.03 & 30.48 & 13.30 \\
\ModelMiniCPMVcite{} & -- & 15.20 & 12.84 & 13.79 & 5.82 \\
\ModelMiniCPMOcite{} & -- & 17.75 & 19.68 & 29.24 & 9.22 \\
\ModelGemmacite{} & -- & 14.08 & 17.49 & 19.67 & 4.45 \\
\ModelSkyworkCite{} & -- & 6.37 & 14.67 & 13.32 & 1.38 \\
\ModelLlamaFullCite{} & -- & 35.77 & 14.38 & 11.34 & 5.33 \\
\ModelQwenSmallcite{} & -- & 30.45 & 16.69 & 26.17 & 7.73 \\
\ModelQwenMediumcite{} & -- & 37.23 & 23.05 & 29.63 & 13.30 \\
\ModelQwenLargecite{} & -- & 29.30 & 19.94 & 27.75 & 10.50 \\
\ModelGPTFourOFullOCite{} & -- & 36.43 & 31.02 & 32.99 & 19.55 \\
\ModelQwenMaxcite{} & -- & 34.79 & 17.83 & 31.41 & 8.91 \\
\ModelGeminiFlashCite{} & -- & 38.89 & 36.76 & 40.03 & 21.47 \\
\midrule
\ModelGPTFiveThinkCite{} & \checkmark & 49.70 & 38.84 & 38.33 & 24.95 \\
\ModelGeminiProThinkCite{} & \checkmark & 52.35 & 44.8 & 43.08 & 29.59 \\
\midrule
\ModelQwenSurgCite{} & -- & 50.30 & 38.40 & 56.75 & 23.47 \\
\midrule
\textbf{Surg-R1 (Ours)} & \checkmark & \textbf{80.90} & \textbf{67.26} & \textbf{71.81} & \textbf{55.87} \\
\bottomrule
\end{tabular}
\label{tab:cholec80_phase_supp}
\end{table*}

\begin{table*}[h!]
\centering
\caption{{\bf Action recognition performance on the SAR-RARP50 dataset.} SAR-RARP50 contains 50 robot-assisted radical prostatectomy procedures with fine-grained surgical action annotations. All metrics are reported in percentage (\%). Best results are \textbf{bolded}.}
\begin{tabular}{llcccc}
\toprule
 & & \multicolumn{4}{c}{RARP Action Recognition (\%)} \\
\cmidrule(lr){3-6}
Model & Thinking & Accuracy & Recall & Precision & Jaccard \\
\midrule
\ModelLLaVAcite{} & -- & 5.10 & 12.50 & 0.64 & 0.64 \\
\ModelPhicite{} & -- & 15.10 & 12.83 & 11.41 & 3.47 \\
\ModelMistralFullCite{} & -- & 12.50 & 12.68 & 5.89 & 1.78 \\
\ModelInternVLSmallcite{} & -- & 29.30 & 12.78 & 9.96 & 6.13 \\
\ModelInternVLLargeCite{} & -- & 29.50 & 12.87 & 22.82 & 6.28 \\
\ModelMiniCPMVcite{} & -- & 24.30 & 12.07 & 13.43 & 7.25 \\
\ModelMiniCPMOcite{} & -- & 30.80 & 13.13 & 20.12 & 5.74 \\
\ModelGemmacite{} & -- & 33.20 & 13.86 & 18.01 & 5.32 \\
\ModelSkyworkCite{} & -- & 12.30 & 13.02 & 19.83 & 3.08 \\
\ModelLlamaFullCite{} & -- & 25.10 & 12.50 & 3.14 & 3.14 \\
\ModelQwenSmallcite{} & -- & 31.10 & 12.50 & 3.89 & 3.89 \\
\ModelQwenMediumcite{} & -- & 31.80 & 13.65 & 41.35 & 5.21 \\
\ModelQwenLargecite{} & -- & 28.20 & 13.04 & 22.30 & 5.94 \\
\ModelGPTFourOFullOCite{} & -- & 28.10 & 15.99 & 16.15 & 9.16 \\
\ModelQwenMaxcite{} & -- & 28.30 & 14.60 & 10.20 & 7.14 \\
\ModelGeminiFlashCite{} & -- & 24.40 & 17.51 & 18.54 & 7.27 \\
\midrule
\ModelGeminiProThinkCite{} & \checkmark & 31.81 & 18.74 & 37.41 & 10.56 \\
\ModelGPTFiveThinkCite{} & \checkmark & 30.64 & 19.42 & 19.90 & 11.21 \\
\midrule
\ModelQwenSurgCite{} & -- & 31.50 & 12.73 & 6.89 & 4.28 \\
\midrule
\textbf{Surg-R1 (Ours)} & \checkmark & \textbf{48.10} & \textbf{29.95} & \textbf{45.49} & \textbf{22.55} \\
\bottomrule
\end{tabular}
\label{tab:rarp_action_supp}
\end{table*}

\begin{table*}[h!]
\centering
\caption{{\bf Surgical phase recognition performance on the MultiBypass140 dataset.} MultiBypass140 contains 140 laparoscopic Roux-en-Y gastric bypass procedures, enabling evaluation of cross-procedure generalization beyond cholecystectomy. All metrics are reported in percentage (\%). Best results are \textbf{bolded}.}
\begin{tabular}{llcccc}
\toprule
 & & \multicolumn{4}{c}{MultiBypass140 Phase Recognition (\%)} \\
\cmidrule(lr){3-6}
Model & Thinking & Accuracy & Recall & Precision & Jaccard \\
\midrule
\ModelLLaVAcite{} & -- & 14.30 & 4.33 & 7.66 & 2.09 \\
\ModelPhicite{} & -- & 6.81 & 6.76 & 7.53 & 3.14 \\
\ModelMistralFullCite{} & -- & 10.78 & 10.44 & 10.83 & 4.05 \\
\ModelInternVLSmallcite{} & -- & 8.29 & 8.02 & 8.65 & 2.46 \\
\ModelMiniCPMVcite{} & -- & 9.08 & 8.83 & 9.23 & 4.06 \\
\ModelMiniCPMOcite{} & -- & 8.85 & 8.94 & 8.14 & 3.86 \\
\ModelGemmacite{} & -- & 8.85 & 8.67 & 6.81 & 1.94 \\
\ModelLlamaFullCite{} & -- & 11.92 & 11.98 & 15.51 & 5.14 \\
\ModelQwenSmallcite{} & -- & 8.29 & 8.17 & 2.53 & 1.72 \\
\midrule
\ModelGeminiProThinkCite{} & \checkmark & 28.72 & 29.19 & 31.74 & 16.61 \\
\ModelGPTFiveThinkCite{} & \checkmark & 29.01 & 29.32 & 27.00 & 15.21 \\
\midrule
\ModelQwenSurgCite{} & -- & 9.76 & 9.78 & 5.46 & 3.36 \\
\midrule
\textbf{Surg-R1 (Ours)} & \checkmark & \textbf{49.94} & \textbf{50.74} & \textbf{55.24} & \textbf{32.02} \\
\bottomrule
\end{tabular}
\label{tab:multibypass_phase_supp}
\end{table*}

\subsection{Detailed Experimental Results on Multi-Center External Validation Datasets}

This section presents performance results on six external validation datasets collected from multiple medical centers. These datasets evaluate the model's generalization capability to real-world clinical environments.

\textbf{Note on Proprietary Model Evaluation:} Due to data privacy regulations and institutional policies, external validation datasets cannot be uploaded to cloud-based APIs. Consequently, proprietary commercial models (GPT-5.1, Gemini 3.0 Pro, etc.) could not be evaluated on these datasets. All evaluations on external validation data were performed using locally deployable models only.

\begin{table}[h]
    \centering
    \small
    \caption{Overview of Multi-Center External Validation Datasets and Associated Surgery Types}
    \label{tab:dataset_surgery_types}
    \begin{tabular}{ll}
        \toprule
        \textbf{Dataset / Institution} & \textbf{Surgery Type} \\
        \midrule
        West China Hospital Dataset & Laparoscopic Cholecystectomy \\
        Nanfang Hospital (SMU) Dataset & Laparoscopic Cholecystectomy \\
        Renji Hospital Dataset & Laparoscopic Cholecystectomy \\
        University Hospital of Strasbourg Dataset & Laparoscopic Cholecystectomy \\
        Chinese University of Hong Kong (CUHK) Dataset & Robot-Assisted Radical Prostatectomy \\
        \bottomrule
    \end{tabular}
\end{table}

\begin{table*}[t]
\centering
\small
\caption{{\bf Critical View of Safety (CVS) assessment on the West China Hospital external validation dataset.} This dataset was collected from laparoscopic cholecystectomy procedures at West China Hospital. Accuracy (\%) is reported; ``Overall'' denotes the accuracy of the joint CVS assessment (all three criteria evaluated together), and ``Avg'' denotes the mean of per-criterion accuracies. Best results are \textbf{bolded}.}
\label{tab:westchina_results}
\begin{tabular}{lccccc}
\toprule
\textbf{Model} & \textbf{Overall} & \textbf{Avg} & \textbf{C1} & \textbf{C2} & \textbf{C3} \\
\midrule
\ModelLLaVAcite{} & 33.52 & 33.41 & 31.08 & 48.35 & 20.81 \\
\ModelPhicite{} & 78.90 & 78.69 & 73.37 & 78.06 & 84.63 \\
\ModelMistralFullCite{} & 71.77 & 71.28 & 59.28 & 84.98 & 69.58 \\
\ModelInternVLSmallcite{} & 46.75 & 47.55 & 67.47 & 30.24 & 44.93 \\
\ModelMiniCPMVcite{} & 38.91 & 39.09 & 43.25 & 21.94 & 52.08 \\
\ModelMiniCPMOcite{} & 39.99 & 40.89 & 63.37 & 22.47 & 36.82 \\
\ModelGemmacite{} & 41.94 & 43.56 & 84.34 & 43.45 & 2.88 \\
\ModelLlamaFullCite{} & 80.01 & 80.17 & 84.34 & 81.36 & 74.81 \\
\ModelQwenSmallcite{} & 54.95 & 56.09 & 84.34 & 31.42 & 52.51 \\
\midrule
\ModelQwenSurgCite{} & 78.94 & 79.16 & 84.34 & 53.14 & \textbf{100.00} \\
\midrule
\textbf{Surg-R1 (Ours)} & \textbf{92.50} & \textbf{92.20} & \textbf{84.82} & \textbf{97.12} & 94.66 \\
\bottomrule
\end{tabular}
\end{table*}

\begin{table*}[t]
\centering
\small
\caption{{\bf Critical View of Safety (CVS) assessment on the Nanfang Hospital (SMU) external validation dataset.} This dataset was collected from laparoscopic cholecystectomy procedures at Nanfang Hospital. Accuracy (\%) is reported; ``Overall'' denotes the accuracy of the joint CVS assessment, and ``Avg'' denotes the mean of per-criterion accuracies. Best results are \textbf{bolded}.}
\label{tab:smu_results}
\begin{tabular}{lccccc}
\toprule
\textbf{Model} & \textbf{Overall} & \textbf{Avg} & \textbf{C1} & \textbf{C2} & \textbf{C3} \\
\midrule
\ModelLLaVAcite{} & 33.33 & 33.33 & 34.48 & 37.93 & 27.59 \\
\ModelPhicite{} & 45.98 & 45.98 & 68.97 & 31.03 & 37.93 \\
\ModelMistralFullCite{} & 64.37 & 64.37 & 72.41 & 62.07 & 58.62 \\
\ModelInternVLSmallcite{} & 52.87 & 52.87 & 82.76 & 24.14 & 51.72 \\
\ModelMiniCPMVcite{} & 36.78 & 36.78 & 37.93 & 17.24 & 55.17 \\
\ModelMiniCPMOcite{} & 40.23 & 40.23 & 68.97 & 10.34 & 41.38 \\
\ModelGemmacite{} & 63.22 & 63.22 & 89.66 & 86.21 & 13.79 \\
\ModelLlamaFullCite{} & 81.61 & 81.61 & 89.66 & 79.31 & 75.86 \\
\ModelQwenSmallcite{} & 48.28 & 48.28 & 89.66 & 13.79 & 41.38 \\
\midrule
\ModelQwenSurgCite{} & 80.46 & 80.46 & 89.66 & 62.07 & \textbf{89.66} \\
\midrule
\textbf{Surg-R1 (Ours)} & \textbf{87.36} & \textbf{87.36} & \textbf{89.66} & \textbf{96.55} & 75.86 \\
\bottomrule
\end{tabular}
\end{table*}

\begin{table*}[t]
\centering
\small
\setlength{\tabcolsep}{5pt}
\caption{{\bf Triplet recognition performance on the Renji Hospital external validation dataset.} This dataset was collected from laparoscopic cholecystectomy procedures at Renji Hospital. mAP is reported for Instrument, Verb, Target, and the complete Triplet. Best results are \textbf{bolded}.}
\label{tab:renji_triplet}
\begin{tabular}{lcccc}
\toprule
\multirow{2}{*}{\textbf{Model}} & \multicolumn{4}{c}{\textbf{mAP (\%)}} \\
\cmidrule(lr){2-5}
 & \textbf{Inst.} & \textbf{Verb} & \textbf{Target} & \textbf{Triplet} \\
\midrule
\ModelLLaVAcite{} & 28.23 & 20.00 & 14.22 & 5.79 \\
\ModelPhicite{} & 27.65 & 20.20 & 13.47 & 5.79 \\
\ModelMistralFullCite{} & 28.98 & 20.45 & 13.68 & 6.05 \\
\ModelInternVLSmallcite{} & 30.54 & 21.29 & 13.28 & 5.92 \\
\ModelMiniCPMVcite{} & 27.50 & 19.87 & 13.08 & 5.79 \\
\ModelMiniCPMOcite{} & 28.45 & 20.31 & 15.07 & 6.43 \\
\ModelGemmacite{} & 29.15 & 19.90 & 13.57 & 5.79 \\
\ModelLlamaCite{} & 36.16 & 25.13 & 14.27 & 6.72 \\
\ModelQwenSmallcite{} & 28.87 & 19.38 & 13.08 & 5.85 \\
\midrule
\ModelQwenSurgCite{} & 35.99 & 22.15 & 15.63 & 7.49 \\
\midrule
\textbf{Surg-R1 (Ours)} & \textbf{51.91} & \textbf{26.65} & \textbf{23.09} & \textbf{10.05} \\
\bottomrule
\end{tabular}
\end{table*}

\begin{table*}[t]
\centering
\small
\setlength{\tabcolsep}{8pt} 
\caption{{\bf Surgical phase recognition performance on the Renji Hospital external validation dataset.} This dataset was collected from laparoscopic cholecystectomy procedures at Renji Hospital. Metrics include Accuracy, Macro-Precision, Macro-Recall, and Macro-Jaccard. All values are in percentages (\%). Best results are \textbf{bolded}.}
\label{tab:renji_phase_detailed}
\begin{tabular}{lcccc}
\toprule
\textbf{Model} & \textbf{Accuracy} & \textbf{Precision} & \textbf{Recall} & \textbf{Jaccard} \\
\midrule
\ModelLlamaFullCite{} & 31.38 & 25.64 & 23.46 & 13.41 \\
\ModelGemmacite{} & 23.83 & 12.64 & 17.41 & 5.75 \\
\ModelMistralFullCite{} & 20.96 & 20.72 & 16.35 & 6.89 \\
\ModelLLaVAcite{} & 20.21 & 11.90 & 8.38 & 3.58 \\
\ModelInternVLSmallcite{} & 17.87 & 19.92 & 18.77 & 8.27 \\
\ModelMiniCPMVcite{} & 14.68 & 24.54 & 16.01 & 4.85 \\
\ModelPhicite{} & 10.85 & 13.04 & 13.73 & 2.94 \\
\ModelQwenSmallcite{}{} & 10.74 & 1.54 & 14.29 & 1.54 \\
\ModelMiniCPMOcite{} & 10.64 & 18.68 & 17.11 & 5.29 \\
\midrule
\ModelQwenSurgCite{} & 20.74 & 36.31 & 27.02 & 10.42 \\
\midrule
\textbf{Surg-R1 (Ours)} & \textbf{57.66} & \textbf{65.36} & \textbf{62.50} & \textbf{41.15} \\
\bottomrule
\end{tabular}
\end{table*}

\begin{table*}[t]
\centering
\small
\setlength{\tabcolsep}{8pt}
\caption{{\bf Surgical phase recognition performance on the University Hospital of Strasbourg external validation dataset.} This dataset was collected from laparoscopic cholecystectomy procedures at the University Hospital of Strasbourg. Metrics include Accuracy, Macro-Precision, Macro-Recall, and Macro-Jaccard. All values are in percentages (\%). Best results are \textbf{bolded}.}
\label{tab:strasbourg_phase_detailed}
\begin{tabular}{lcccc}
\toprule
\textbf{Model} & \textbf{Accuracy} & \textbf{Precision} & \textbf{Recall} & \textbf{Jaccard} \\
\midrule
\ModelLlamaFullCite{} & 40.33 & 35.68 & 24.53 & 16.63 \\
\ModelLLaVAcite{} & 39.50 & 19.46 & 9.90 & 6.88 \\
\ModelInternVLSmallcite{} & 37.30 & 20.68 & 15.38 & 9.68 \\
\ModelPhicite{} & 35.50 & 19.70 & 16.21 & 9.90 \\
\ModelGemmacite{} & 35.33 & 25.84 & 19.46 & 10.86 \\
\ModelQwenSmallcite{} & 31.03 & 40.26 & 19.97 & 8.92 \\
\ModelMiniCPMOcite{} & 26.93 & 16.15 & 19.75 & 8.52 \\
\ModelMistralFullCite{} & 25.87 & 19.76 & 15.33 & 8.76 \\
\ModelMiniCPMVcite{} & 24.33 & 15.22 & 16.66 & 7.90 \\
\midrule
\ModelQwenSurgCite{} & 62.57 & 51.94 & 40.87 & 25.05 \\
\midrule
\textbf{Surg-R1 (Ours)} & \textbf{79.80} & \textbf{69.22} & \textbf{61.48} & \textbf{47.76} \\
\bottomrule
\end{tabular}
\end{table*}

\begin{table*}[t]
\centering
\small
\setlength{\tabcolsep}{8pt}
\caption{{\bf Action recognition performance on the Chinese University of Hong Kong (CUHK) external validation dataset.} This dataset was collected from robot-assisted radical prostatectomy procedures. Metrics include Accuracy, Precision, Recall, and Jaccard. All values are in percentages (\%). Best results are \textbf{bolded}.}
\label{tab:surgical_instruction_detailed}
\begin{tabular}{lcccc}
\toprule
\textbf{Model} & \textbf{Accuracy} & \textbf{Precision} & \textbf{Recall} & \textbf{Jaccard} \\
\midrule
\ModelLLaVAcite{} & 24.81 & 17.40 & 11.79 & 4.38 \\
\ModelQwenSmallcite{} & 18.96 & 5.98 & 18.96 & 4.47 \\
\ModelMiniCPMOcite{} & 18.68 & 16.18 & 18.11 & 8.05 \\
\ModelMistralFullCite{} & 18.40 & 13.15 & 18.40 & 7.82 \\
\ModelPhicite{} & 18.30 & 12.57 & 18.21 & 5.97 \\
\ModelGemmacite{} & 18.11 & 8.45 & 18.11 & 4.51 \\
\ModelMiniCPMVcite{} & 17.08 & 14.11 & 17.08 & 7.68 \\
\ModelInternVLSmallcite{} & 16.89 & 14.55 & 16.70 & 7.21 \\
\ModelLlamaFullCite{} & 14.62 & 10.62 & 14.62 & 3.54 \\
\midrule
\ModelQwenSurgCite{} & 19.15 & 11.01 & 19.15 & 4.96 \\
\midrule
\textbf{Surg-R1 (Ours)} & \textbf{32.92} & \textbf{38.06} & \textbf{32.92} & \textbf{18.97} \\
\bottomrule
\end{tabular}
\end{table*}